\pdfoutput=1

\documentclass[11pt]{article}

\usepackage[]{acl}

\usepackage{times}
\usepackage{latexsym}

\usepackage{graphicx}
\usepackage{scalefnt}
\usepackage{svg}
\usepackage{comment}
\usepackage{booktabs}
\usepackage{array}
\usepackage{caption}
\usepackage{subcaption}
\usepackage{tablefootnote}
\usepackage{multirow}
\usepackage{multicol}
\usepackage{enumitem}
\usepackage{xcolor, color, soul}
\usepackage[multiple]{footmisc}
\usepackage{amsmath}
\usepackage{csquotes}
\usepackage{fixltx2e}
\usepackage{hyperref}
\usepackage{pifont}
\usepackage{makecell}
\usepackage{longtable}
\usepackage{tabularx}
\usepackage{supertabular}

\newcolumntype{L}[1]{>{\raggedright\arraybackslash}m{#1}}
\newcolumntype{R}[1]{>{\raggedleft\arraybackslash}m{#1}}
\newcolumntype{C}[1]{>{\centering\arraybackslash}m{#1}}

\usepackage{listings}
\usepackage{tcolorbox}

\usepackage[T1]{fontenc}

\usepackage[utf8]{inputenc}

\usepackage{microtype}

\usepackage{inconsolata}

\newcommand\method{\textsc{InstrucTE}}
\newcommand\task{\textsc{Schema-to-Json}}

\newcommand\axcell{\textsc{AxCell}}

\newcommand{\discomat}{\textsc{DisCoMat}}

\newcolumntype{C}[1]{>{\PreserveBackslash\centering}p{#1}}

\usepackage[normalem]{ulem}

\title{Schema-Driven Information Extraction from Heterogeneous Tables}

\author{Fan Bai$^{\clubsuit}$ Junmo Kang$^{\clubsuit}$  Gabriel Stanovsky$^{\diamondsuit}$  Dayne Freitag$^{\spadesuit}$ 
\\ \textbf{Mark Dredze}$^{\heartsuit}$ \textbf{Alan Ritter}$^{\clubsuit}$ \\
  $\clubsuit$ {College of Computing, Georgia Institute of Technology} \\
  $\diamondsuit$ {School of Computer Science and Engineering, The Hebrew University of Jerusalem} \\
  $\spadesuit$ \text{Artificial Intelligence Center, SRI International} \\
  $\heartsuit$ \text{Department of Computer Science, Johns Hopkins University} \\
  \small
 \texttt{\{fan.bai, alan.ritter\}@cc.gatech.edu}, \texttt{junmo.kang@gatech.edu}, \\
\small
\texttt{gabriel.stanovsky@mail.huji.ac.il}, \texttt{daynefreitag@sri.com}, \texttt{mdredze@cs.jhu.edu} \\
 \\
 }

\begin{document}
\maketitle
\begin{abstract}

In this paper, we explore the question of whether large language models can support cost-efficient information extraction from tables. We introduce {\em schema-driven information extraction}, a new task that transforms tabular data into structured records following a human-authored schema. To assess various LLM's capabilities on this task, we present a benchmark comprised of tables from four diverse domains: machine learning papers, chemistry literature, material science journals, and webpages. We use this collection of annotated tables to evaluate the ability of open-source and API-based language models to extract information from tables covering diverse domains and data formats. Our experiments demonstrate that surprisingly competitive performance can be achieved without requiring task-specific pipelines or labels, achieving F\textsubscript{1} scores ranging from 74.2 to 96.1, while maintaining cost efficiency. Moreover, through detailed ablation studies and analyses, we investigate the factors contributing to model success and validate the practicality of distilling compact models to reduce API reliance.\footnote{Our code and data are available at \url{https://github.com/bflashcp3f/schema-to-json}.}

\end{abstract} 

\section{Introduction}
\label{sec:intro}

\begin{figure}[!t]
\begin{center}
  \includegraphics[width=0.40\textwidth]
  {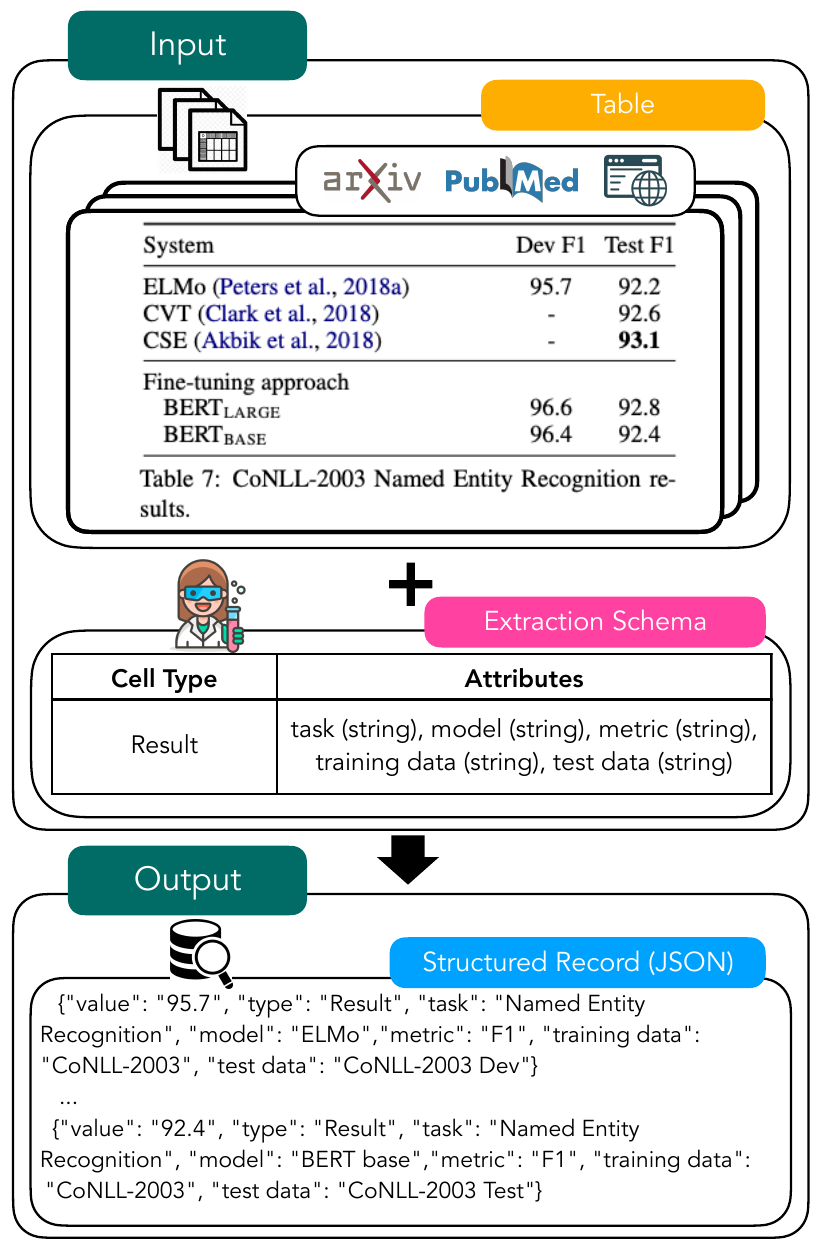}
\caption{Overview of Schema-Driven Information Extraction. The input includes two elements: the source code of a table and a human-authored extraction schema, outlining the target attributes and their data types. The output consists of a sequence of JSON records that conform to the extraction schema. 
}
  \label{fig:task}
\end{center}
\end{figure}

Vast quantities of experimental data are locked away in tables found in scientific literature.
These tables are primarily designed for visual presentation, and the underlying data is typically not available in any structured format, such as a relational or graph database. Some table collections have simple or uniform structures \cite{cafarella2008webtables}, making them easy to convert to relational data, for example, Wikipedia tables \cite{lebret2016neural,iyyer2017search}, however a lot of information is stored in tables with complex and varied layouts, such as tables of results in papers found on {\tt arXiv.org}.

Prior work on extracting data from tables has focused on developing custom pipelines for each new table format or domain, for example extracting machine learning leaderboards from \LaTeX~result tables \cite{kardas-etal-2020-axcell}. 
Importantly, the development of these specialized pipelines necessitates domain-specific labeled data, which not only incurs a significant cost in collection for every new extraction task but also constrains their applicability outside the originating domain.

In this paper, we show how LLMs can enable accurate domain-independent extraction of data from heterogeneous tables.
We present a new formulation of the table extraction problem, which we refer to as Schema-Driven Information Extraction.
In Schema-Driven IE, the only human supervision provided is a schema that describes the data model, including the target attributes and their data types, formulated in a JSON format.\footnote{JSON is chosen as the output format for two main reasons: 1) its widespread use ensures a significant representation in the LLM's pre-training corpus, which is crucial for optimizing model performance; and 2) its simplicity in parsing and processing, especially its support for one-line output, makes it advantageous for outputs spanning multiple cells, offering a clear benefit over indent-based formats like YAML.}
Given an extraction schema, and a table as input, the model then outputs a sequence of JSON objects, each of which describes a table cell.
For example, as demonstrated in Figure \ref{fig:task}, a domain expert outlines the attributes of interest related to result cells in a machine learning table, and the model extracts JSON objects following this schema.

To evaluate the ability of LLMs to perform Schema-Driven IE, we introduce a new benchmark consisting of table extraction datasets in four diverse domains: machine learning papers, chemistry literature, material science journals, and webpages - each of which has a different data format (\LaTeX, XML, CSV, and HTML, respectively). We curate and annotate new datasets for the first two domains, while adapting existing datasets for the latter two. 

Using this newly developed benchmark, we analyze the performance of open-source and proprietary LLMs. We find that state-of-the-art proprietary models are capable of accurately extracting data from diverse domains and table formats without supervision.  For example, GPT-4 \cite{gpt4} and {\tt code-davinci} \cite{codex}, are capable of accurate table extraction (ranging from 74.2 to 96.1 F\textsubscript{1}), given only a relevant data schema as input to define the task. This performance is comparable to fully supervised models, which operate at an F\textsubscript{1} range of about 64.1 to 96.1. We also present a number of analyses on various factors that are key to achieving good performance while minimizing inference costs, including retrieving text from outside the table, in addition to an iterative error recovery strategy.
Moreover, we demonstrate the utility of Schema-Driven IE by evaluating performance on the downstream task of leaderboard extraction from machine learning papers \citep{kardas-etal-2020-axcell}.




\begin{figure*}[ht!]
    \centering
    \includegraphics[width=\textwidth]{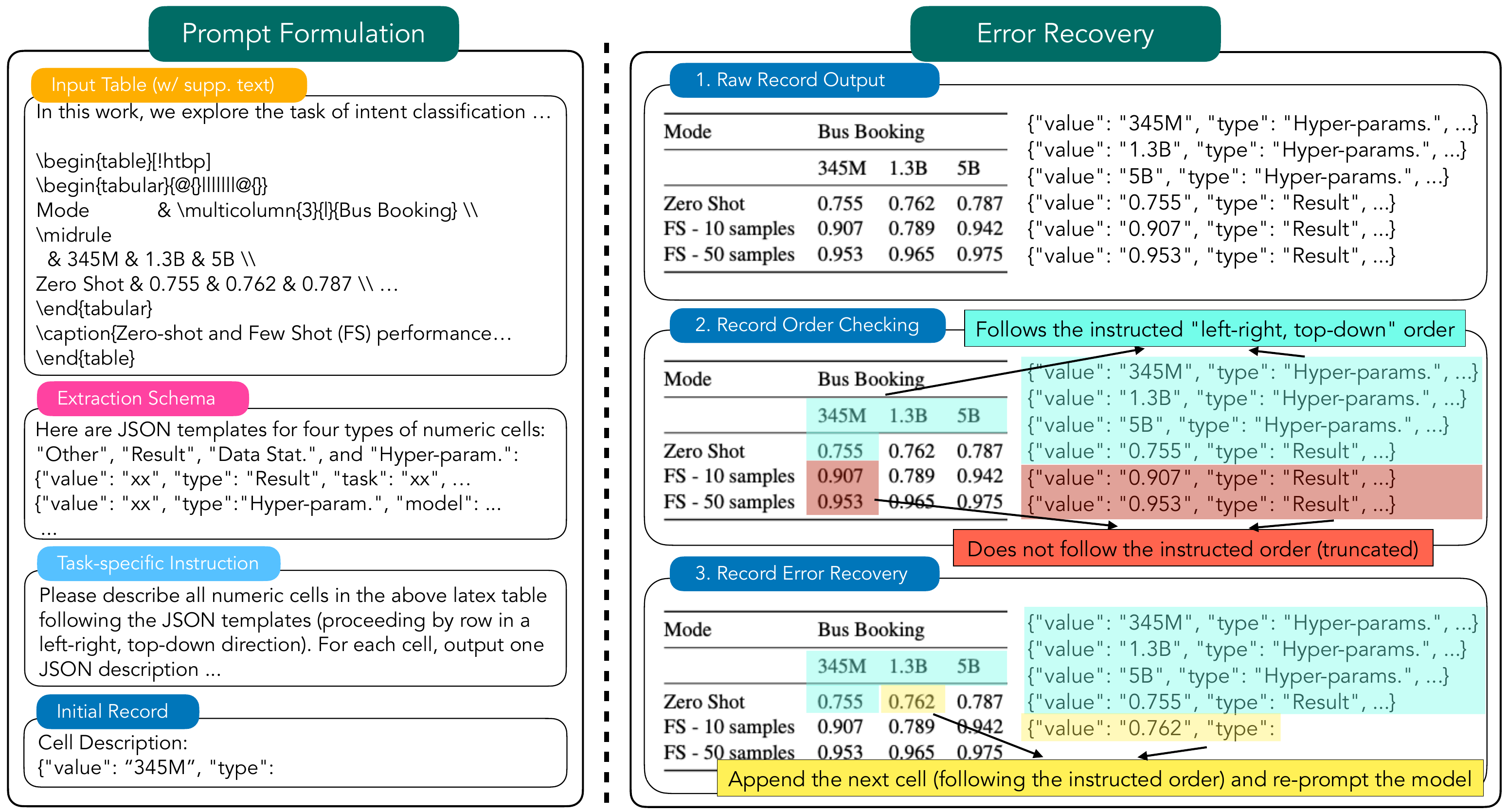}
    \caption{Left: Prompt formulation of our proposed method \method{}. Right: Illustration of our error-recovery strategy, which ensures the model compliance of the instructed cell traversal order and reduces inference costs.
    }
    \label{fig:method}
\end{figure*}

\section{Schema-Driven Information Extraction}
\label{sec:task_def}



We now describe Schema-Driven IE, a new task that extracts structured records from tables with minimal supervision. As shown in Figure \ref{fig:task}, the task input contains two elements: 1) a table with numerous cells, optionally supplemented with contextual text, e.g., retrieved paragraphs from the same document;  and 2) an extraction schema that outlines target attributes and their data types for various record types (implemented as JSON templates). Given the input, the model generates a sequence of JSON objects, where each object corresponds to a cell in the table and contains key-value pairs for the pre-defined attributes of a specific record type.

Consider a table in an ML paper that displays various models' results.
Our proposed task enables the extraction of result records from each cell in the table.  These records include relevant attributes such as the evaluation metric, task, etc, which are structured in corresponding JSON objects and could facilitate meta-analysis of experiments or support research on reproducibility.

To demonstrate the feasibility of Schema-Driven IE on tables, we introduce \textbf{\method{}}, a method to extract structured records from a broad range of semi-structured data, using only task-specific instructions. \method{} uses a template-based approach to information extraction \citep{chambers2011template,chen2023unified}, where the extraction schema is represented as a series of JSON templates. The underlying LLM is instructed to select the appropriate template and populate it with extracted values for each cell in an input table, following a specified cell traversal order. As illustrated in Figure \ref{fig:method} (left), the prompt used by \method{} consists of four key components: an input table (optionally) supplemented with contextual text, an extraction schema, task-specific instructions, and an initial record for starting the process. 

Despite explicit instructions, we found that models often fail to generate JSON records for all the cells in a single inference pass.  Instead, models often deviate from the instructed cell traversal order, leading to partial extraction of the input table's cells.
To mitigate this, we use an iterative error recovery strategy. As shown on the right side of Figure \ref{fig:method}, we detect deviations from the instructed \textit{left-right, top-down} order by comparing predicted cell values with those from a rule-based cell detector. 
Then, we truncate the LLM’s output to the point of deviation, and re-prompt the model with the truncated sequence, adding the value of the next target cell. This process is repeated until all records are generated. 
Using identified cells as a scaffold, this strategy helps the model adhere to the instructed order, significantly improving performance despite potential propagated errors in cell identification.
In Section \ref{sec:ablation}, we show that our approach is much more cost-efficient than cell-by-cell prompting while achieving similar performance.
For more details on \method{}, including prompt formulation and cell detectors, please refer to Appendix \ref{sec:promp_appendix}.

\section{The \task{} Benchmark}
\label{sec:eval}

We now present the details of our benchmark, \task{}, which is designed to assess the capabilities of LLMs to extract data from tables, adhering to a predefined schema. This benchmark contains tables from four domains: machine learning papers, chemistry literature, materials science journals, and webpages. Each domain features a unique textual format, namely, \LaTeX, XML, CSV, and HTML, requiring models to generalize across domains and formats. For ML tables, we add relevant paragraphs from the same documents to provide additional context, testing the models' capacity to jointly understand tabular and textual data.
We manually annotate datasets for the first two domains and adapt pre-existing datasets into our unified format for the latter two. Statistics of the four datasets are summarized in Table \ref{tab:dataset_stat}.

\paragraph{arXiv Machine Learning Tables} We create a manually annotated dataset focused on tables from arXiv ML papers, emphasizing numeric cells that are classified into four categories: Results, Hyper-parameters, Data Statistics, or Other. Extraction attributes are pre-defined for the first three categories; for instance, result records incorporate textual attributes such as \textit{evaluation metric} (e.g., F\textsubscript{1}) and \textit{dataset} (e.g., SQuAD), as shown in Figure~\ref{fig:task}. 
To avoid data contamination with top models like GPT-4 (\texttt{0613}),\footnote{According to OpenAI \href{https://platform.openai.com/docs/models/gpt-4-turbo-and-gpt-4}{website}, GPT-4 (\texttt{0613}) was trained on data until Sep. 2021.} we collected papers published after the knowledge cutoff (between October and November 2022) from three subfields: Machine Learning, Computer Vision, and Natural Language Processing. Five tables were randomly selected from each paper, including appendices.
We employ computer scientists with ML backgrounds for annotation, and evaluate inter-annotator agreement (IAA) score by calculating F\textsubscript{1} (see Section \ref{sec:eval_metric} for details) on double-annotated tables, treating one set of annotations as gold labels and the other as predictions. This method yields an F\textsubscript{1} score of 96.6 when applying thresholded token-level F\textsubscript{1} for attribute matching.  For additional information on ML tables, including predefined attributes and the annotation process, please refer to Appendix \ref{sec:mltab_appendix}.

\newcommand{\subfifty}[1]{$\text{#1}_{50}$}
\newcommand{\icfifty}{\subfifty{IC}}

\begin{table}[tb!]
\begin{center}
\scalebox{0.73}{
\begin{tabular}{lrrrr}
\toprule
   & \textbf{ML}  & \textbf{Chemistry} & \textbf{\discomat{}} & \textbf{\textsc{SWDE}}  \\

 & {(ours)}  & {(ours)} & {(\citeyear{gupta2022discomat})} & (\citeyear{swde}) \\
  \midrule
\normalsize{Textual format} & \normalsize{\LaTeX} & \normalsize{XML} & \normalsize{CSV} & \normalsize{HTML} \\
\normalsize{\# cell types} & \normalsize{4} & \normalsize{6} & \normalsize{2} & \normalsize{8} \\
\normalsize{\# attr. types} & \normalsize{11} & \normalsize{4} & \normalsize{4} & \normalsize{32} \\
\normalsize{\# papers (web.)} & \normalsize{25} & \normalsize{16} & \normalsize{656} & \normalsize{80} \\ 
\normalsize{\# tables (pages)} & \normalsize{122} & \normalsize{26} & \normalsize{1,031} & \normalsize{1,600}  \\
\normalsize{\# anno. records} & \normalsize{3,792} & \normalsize{1,498} & \normalsize{9,036} & \normalsize{1,600} \\
\normalsize{\# records / table} & \normalsize{31.1} & \normalsize{57.6} & \normalsize{8.8} & \normalsize{1} \\
\bottomrule
\end{tabular}
}
\end{center}
\caption{\label{tab:dataset_stat} Dataset statistics of four datasets in our \task{} benchmark. 
}
\end{table}

\paragraph{PubMed Chemistry Tables} We also annotate a new dataset of PubMed tables describing the physical properties of chemical compounds. The automated extraction of physical properties from such tables could provide substantial real-world benefits, for example collecting much-needed data for training ML models that can support inverse molecular design \citep{kim2018deep} and thus accelerating the drug design process \citep{Fields19,STOKES2020688}. Here, we focus on cells concerning five important physical properties identified by chemists: \icfifty{}, \subfifty{EC}, \subfifty{GI}, \subfifty{CC}, and MIC.\footnote{\url{https://www.sciencedirect.com/topics/pharmacology-toxicology-and-pharmaceutical-science/ic50}} Three common attributes are manually extracted from tables for all properties: \emph{unit}, \emph{treatment} (experimental compound), and \emph{target} (measured biological entity, e.g., a gene expression). Similar to the ML tables, domain experts annotate JSON records for relevant cells, and Table-F\textsubscript{1} calculated on double-annotated tables is used as the IAA score. A Table-F\textsubscript{1} score of 91.0 underscores the reliability of the dataset.

\paragraph{\discomat{} \citep{gupta2022discomat}} 
We experiment with \discomat{}, a dataset focusing on glass composition tables from Elsevier material science journals. The task is to extract tuples comprising (\textit{material}, \textit{constituent}, \textit{percentage}, \textit{unit}) from given tables. We adapt \discomat{} to fit our Schema-Driven IE framework by grounding the \textit{percentage} element to numeric cells in the table and considering the other elements as attributes. The model is tasked to identify numeric cells representing constituent percentages and predict the associated three attributes. 
We refer readers to \citet{gupta2022discomat} for more details of \discomat{}.\footnote{In the released corpus, tables are represented as matrices; we, therefore, transform them into CSV tables (using the pipe symbol "|" as the delimiter) prior to feeding them into LLMs.}

\paragraph{\textsc{SWDE} \citep{swde}} Finally, we add SWDE (Structured Web Data Extraction) as a fourth dataset, aimed at extracting pre-defined attributes from HTML webpages. This dataset comprises roughly 124K pages gathered from eight distinct verticals, such as \textit{Auto}, \textit{Book}, and \textit{Movie}. Each vertical includes ten unique websites and is associated with a set of 3 to 5 target attributes. For instance, the \textit{Movie} vertical seeks to extract attributes such as \texttt{title}, \texttt{director}, and \texttt{genre}.


\section{Experiments}
\label{sec:exp}

\begin{figure*}[ht!]
    \centering
    \includegraphics[width=0.9\textwidth]{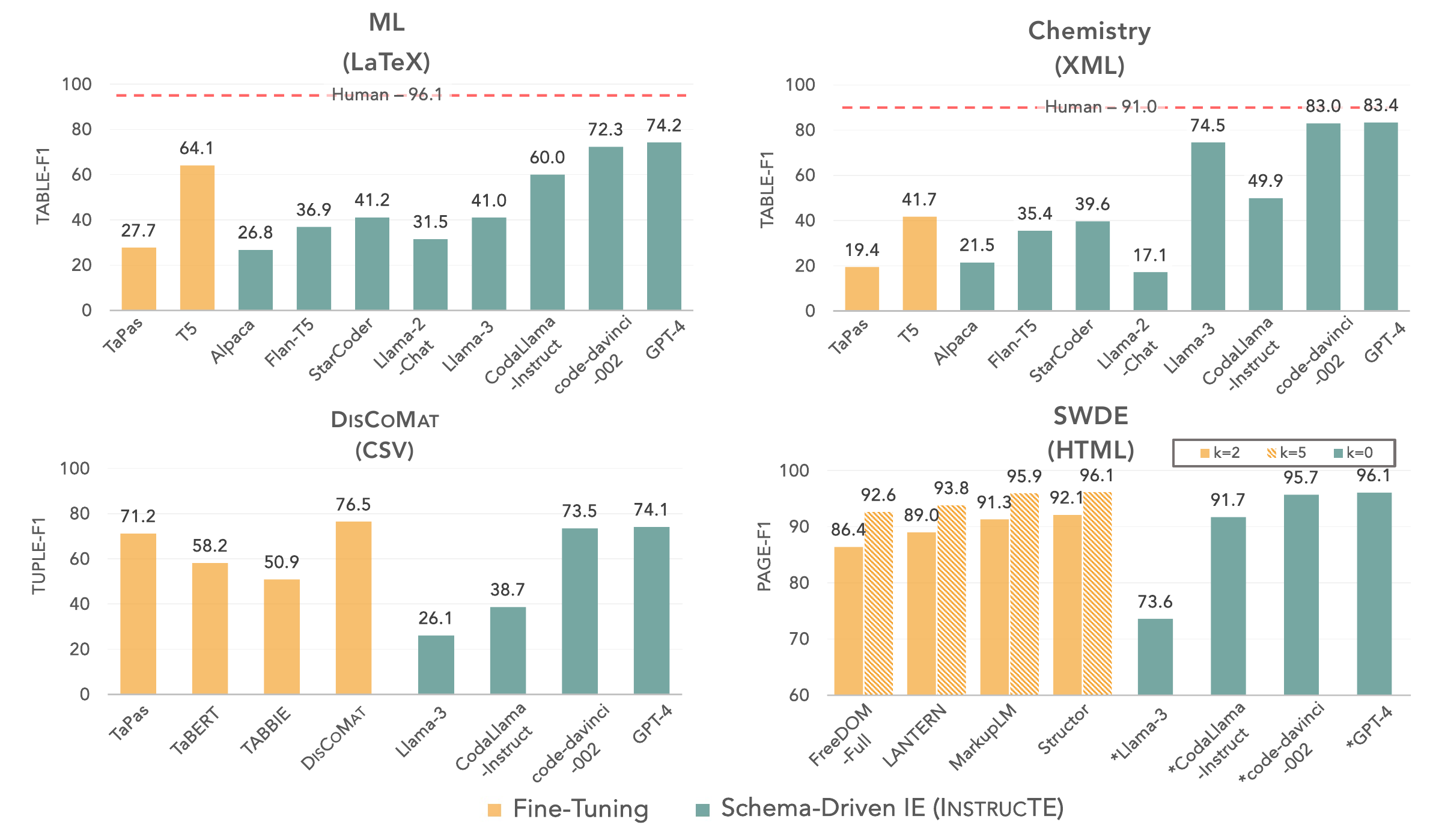}
    \caption{Capability of various LLMs to perform Schema-Driven IE, measured using the \task{} benchmark. 
    We employ Table-F\textsubscript{1} for our two newly annotated datasets and provide a measure of human performance.
    For \discomat{} \citep{gupta2022discomat} and SWDE \citep{swde}, we adhere to their original evaluation metrics, i.e., Tuple-F\textsubscript{1} and Page-F\textsubscript{1} respectively, to support comparisons with established methods. In SWDE experiments, $\textit{k}$ represents the number of trained websites from each vertical. 
    Due to API cost constraints, *\method{}'s results are computed on a 1,600 webpage sample, with bootstrap confidence intervals calculated to validate the reliability of these performance estimates (margin of error for 95\% confidence interval with 1000 samples is 0.00995.)
    }
    \label{fig:main_results}
\end{figure*}

We evaluate the capability of various LLMs to perform Schema-Driven IE, in addition to full fine-tuning using our benchmark.
For ML and chemistry tables, we use a subset of 10 and 7 randomly sampled papers separately for model development, which facilitates the training of supervised baselines. For the two pre-existing datasets, we follow the data splits used in the original experiments.

\subsection{Evaluation}
\label{sec:eval_metric}
To evaluate predicted JSON records, we report Table-F\textsubscript{1}, a reference-based metric gauging attribute prediction performance within a table. Table-F\textsubscript{1} represents the harmonic mean of precision and recall, with precision being the ratio of correctly predicted attributes to total predicted attributes.
At the attribute level, we report results using exact match (EM), in addition to a threshold-based token-level similarity.
The threshold is tuned on dev data to maximize alignment between our estimated model performance and performance measured using human judgments (see Appendix \ref{sec:eval_metric_app} for more details). 
We macro-average Table-F\textsubscript{1}, given the wide variance in table sizes.

For \discomat{} and SWDE, we use similar metrics specified in their original papers to support comparison with prior work. We report Tuple-F\textsubscript{1} (\citealp{gupta2022discomat}) for \discomat{}, where a predicted 4-element tuple is considered correct only if it exactly matches the gold tuple. For SWDE,  we report Page-F\textsubscript{1} \citep{swde}, which measures the number of pages where the attributes are accurately predicted.\footnote{Notably, SWDE primarily focuses on identifying textual HTML nodes containing attribute values rather than exact text spans, so we use token-level F\textsubscript{1} to identify the most relevant HTML node for each extracted attribute.} 

To further validate our conclusions, we also present the results of full human evaluation of model outputs in \S \ref{sec:performance_analysis}.

\subsection{Baselines \& Implementation Details}
\label{sec:baseline}

We evaluate the capability of multiple LLMs to perform Schema-Driven IE, including API-based GPT-4 and GPT-3.5 models and open-source models, such as Llama3-8B-Instruct \cite{llama3modelcard}, Llama2-Chat-13B \citep{touvron2023llama}, CodeLlama-instruct-13B \citep{rozière2023code}, StarCoder-15.5B \citep{starcoder}, LLaMA-7B \citep{touvron2023llama1}, and Alpaca-7B \citep{alpaca}.
We also frame Schema-Driven IE as a TableQA problem, applying multi-choice and extractive QA prompts for template selection and cell attribute prediction, respectively.
Furthermore, we also evaluate T5-11B \citep{t5} and TaPas \citep{herzig-etal-2020-tapas}, a table-specialized LM. For implementation details of \method{} and other methods, see Appendix \ref{sec:implement_details}.\footnote{We developed a rule-based method for chemistry tables based on the training set, which only achieved a Table-F\textsubscript{1} score of 51.3, significantly lower than our proposed InstrucTE. Due to the substantial effort required to create specialized rule-based systems for each domain and the performance gap, we decided not to pursue this approach further.}

For \discomat{} and SWDE, we compare \method{} with established baselines, which either design task-specific architectures, such as FreeDom \citep{freedom} and LANTERN \citep{lantern}, or use LMs pretrained on tables or web pages, like TaPas \citep{herzig-etal-2020-tapas}, TaBERT \cite{yin2020tabert}, and MarkupLM \citep{li-etal-2022-markuplm}.

\begin{figure*}[t!]
    \centering
    \includegraphics[width=0.9\textwidth]{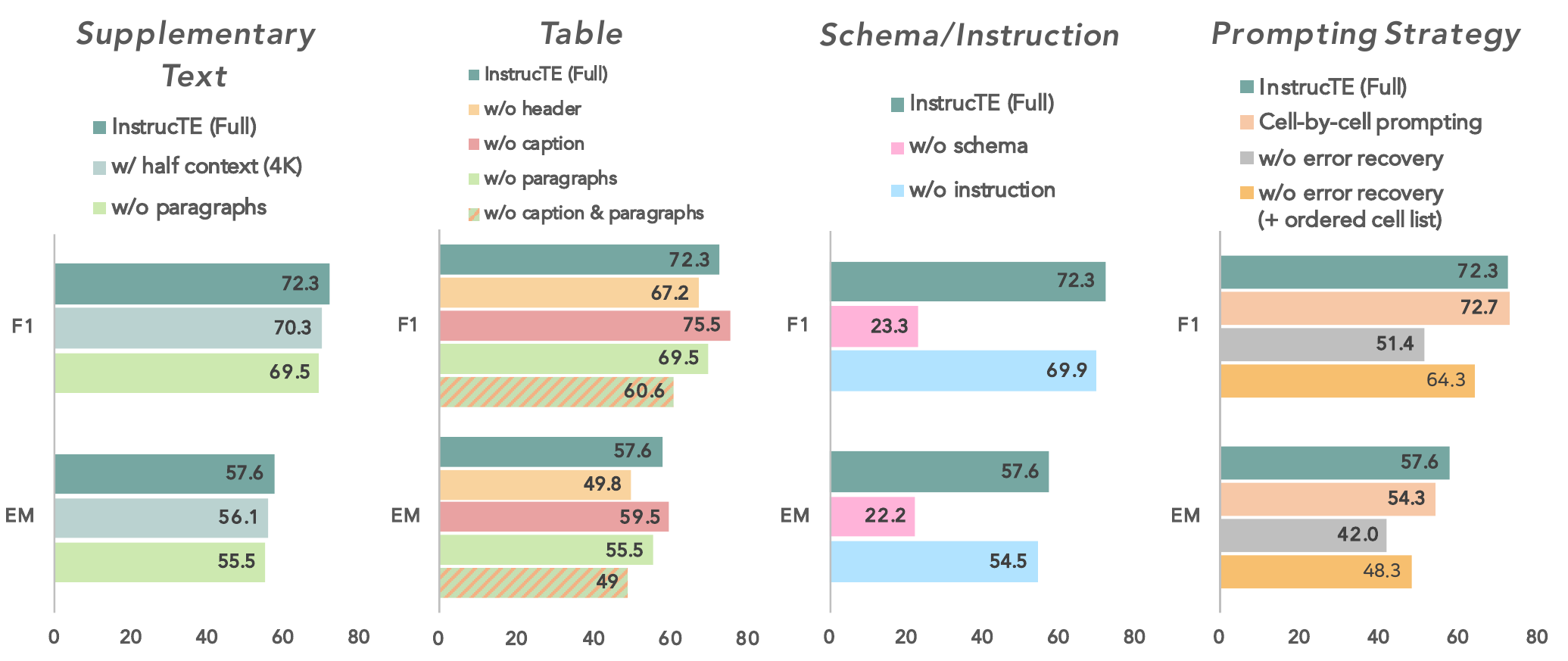}
    \caption{Ablation studies on various components of our \textsc{InstrucTE} (w/ \texttt{code-davinci-002}) on the ML tables. 
    Interestingly, excluding the table caption improves performance. Our detailed analysis in Appendix \ref{sec:error_analysis_of_caption} reveals that low-quality captions (e.g., lack of specificity) may confuse the model, leading to inaccurate predictions.}
    \label{fig:ablation}
\end{figure*}

\subsection{Main Results}
\label{sec:main_results}

Figure \ref{fig:main_results} presents the main results from the comparison between \method{} and other methods on our \task{} benchmark.
We observe that \method{}, in conjunction with API-based models, achieves strong performance across domains and input formats, without any domain-specific labels. With GPT-4, \method{} can outperform fine-tuned models on ML and chemistry tables. However, a substantial disparity remains compared to human performance, e.g., the Table-F\textsubscript{1} on double-annotated examples for ML tables stands at 96.6 when applying thresholded token-level F\textsubscript{1} for attribute matching, which is 22.4 F\textsubscript{1} points higher than GPT-4. 

For \discomat{} and SWDE, GPT-4 performs on par or slightly trails behind the fully supervised state-of-the-art methods, signifying the potential of LLMs to act as flexible, powerful tools for extracting information from tables across diverse data formats and domains.

Despite a noticeable gap compared to API-based LLMs, open-source models show promising results across several domains. For example, CodeLlama-instruct-13B achieves 60.0 Table-F\textsubscript{1} and 91.7 Page-F\textsubscript{1} on ML tables and SWDE, respectively. Llama3-8B achieves 74.5 Table-F\textsubscript{1} on chemistry tables.


\subsection{Ablation Studies}
\label{sec:ablation}

We assess the impact of different components of \textsc{InstrucTE}, including task formulation and error recovery, using ML tables.

\paragraph{LLMs \& Task Formulation} 
In Table \ref{tab:test}, we compare different LLMs, leading to two principal observations. First, code models show strong performance on Schema-Driven IE. This is evident from several key comparisons, such as the performance similarity between \texttt{code-davinci-002} and GPT-4, the superior performance of \texttt{code-davinci-002} compared to other GPT-3.5 models, and the fact that CodeLlama-instruct-13B significantly outperforms Llama2-chat-13B, approaching the performance of \texttt{gpt-3.5-turbo}. This superiority of code models might be attributed to their alignment with Schema-Driven IE, which involves converting table source code into JSON records. Second, non-code open-source models with similar sizes (for instance, those in the 6-7B range) tend to achieve comparable fine-tuning performance, though they might exhibit variations in prompting performance. 

Subsequently, we compare three task formulations: \task{}, TableQA, and Function Calling, which is a feature provided by the OpenAI API.\footnote{\url{https://platform.openai.com/docs/guides/function-calling}} In Function Calling, the schema is formatted as function definitions with attributes serving as arguments. The LM is then tasked with selecting the function and generating JSON objects for extracted arguments on a cell-by-cell basis. From the T5-11B fine-tuning experiments, we observe that \task{} attains better performance than TableQA, demonstrating the value of integrating task-specific instructions and extraction schema in the input. Function Calling with \texttt{gpt-3.5-turbo} shows limited effectiveness, and error analysis suggests that this shortfall primarily stems from the model's struggle in selecting the correct function.\footnote{This finding is supported by a marked performance increase to 63.8 Table-F\textsubscript{1} when the gold function is pre-specified. As each function call yields only one JSON object, this method requires cell-by-cell prompting, which is cost-intensive with GPT-4. Due to API budget constraints, our experiments are limited to \texttt{gpt-3.5-turbo}.}

\begin{table}[!t]
\centering
\resizebox{1.0\linewidth}{!}{
\begin{tabular}{cclcc}
\toprule
\textbf{Exp. Setup} & \textbf{Formulation} & \textbf{Model} & \textbf{Token-F\textsubscript{1}} & \textbf{EM} \\
\midrule

\multirow{6}{*}{\makecell{Fine-tuning \\ (\# Train=1169)}}
& \multirow{2}{*}{TableQA} & TaPas (large) & 27.7 & 21.6 \\ 
&          & T5 (11B) & 61.2 & 46.2 \\ 
  \cmidrule(l){2-5}
& \multirow{4}{*}{\textsc{Sche2JSON}}
& GPT-J (6B) & 49.6 & 38.4 \\
& & LLaMA (7B) & 51.3 & 38.0 \\
& & Alpaca (7B) & 50.2 & 39.4 \\
& & T5 (11B) & 64.1 & 50.2 \\ 

\midrule

\multirow{13}{*}{\makecell{No Fine-tuning}} & TableQA & Flan-T5 (11B) & 36.9 & 27.7 \\ 
 \cmidrule(l){2-5}

& \multirow{1}{*}{Func. Calling}
& \texttt{gpt-3.5-turbo (0613)} & 22.4 & 18.4 \\
\cmidrule(l){2-5}

& \multirow{10}{*}{\textsc{Sche2JSON}}
& GPT-J (6B) & 18.6 & 16.2 \\
& & LLaMA (7B) & 13.5 & 11.5 \\
& & Alpaca (7B) & 26.8 & 21.1 \\
& & Llama2-chat (13B) & 31.5 & 23.0 \\
& & Llama3-instruct (8B) & 41.0 & 32.4 \\
& & StarCoder (15.5B) & 41.2 & 32.3 \\
& & CodeLlama-instruct (13B) & 60.0 & 44.0 \\
& & \texttt{gpt-3.5-turbo (0613)} & 64.1 & 47.9 \\
& & \texttt{text-davinci-003} & 67.4 & 50.4 \\
& & \texttt{code-davinci-002} & 72.3 & 57.6 \\
& & \texttt{gpt-4 (0613)} & \textbf{74.2} & \textbf{58.1} \\

\bottomrule
\end{tabular}
}
\caption{
    \textsc{Test} set performance on ML tables with different LLMs and task formulations. 
   }
\label{tab:test}
\end{table}

\paragraph{Prompt Components \& Error Recovery}
Figure \ref{fig:ablation} shows \method{}'s performance subject to the exclusion of varying prompt components. We use \texttt{code-davinci-002} for these experiments considering API budget limitations and its resemblance to GPT-4 in terms of performance and context length. We observe that removing supplementary text degrades performance.  Table headers contribute positively as expected, while captions surprisingly do not. Further analysis on table captions is provided in Appendix \ref{sec:error_analysis_of_caption}, which suggests that unclear captions can sometimes mislead the model, resulting in inaccurate predictions. Notably, discarding the extraction schema, specifically JSON templates, causes a substantial performance decline, primarily due to attribute name mismatches in the evaluation. Lastly, we show that \method{}'s performance drops significantly without error recovery. Compared to cell-by-cell prompting, error recovery offers similar performance at a fraction of the API cost (\$100 v.s. \$670 on Azure).\footnote{The pricing for \texttt{code-davinci-002} on \href{https://azure.microsoft.com/en-us/pricing/details/cognitive-services/openai-service/}{Azure} is \$0.1 per 1,000 tokens as of June 23rd, 2023.} 

\subsection{Performance Analysis}
\label{sec:performance_analysis}

To further verify our main conclusions from automatic evaluation and gain deeper insights into \method{}’s performance, we conduct a human evaluation and discuss a set of key questions.

\begin{table*}[!ht]
\centering
\resizebox{1\textwidth}{!}{
\begin{tabular}{lclp{10cm}}
\toprule
\textbf{Category} & \textbf{\% (\#)} & \textbf{Fine-grained Error Types} & \textbf{Top 3 Affected Attributes} \\
\midrule

& 31.2 (48) & gold answer in table, not predicted & "Result:experimental\_settings" (39.6\%), "Result:training\_data" (39.6\%), "Result:model\_settings" (20.8\%) \\
False Negative & 14.9 (23) & gold answer in main text, not predicted & "Result:training\_data" (39.1\%), "Hyper-parameter:model" (26.1\%), "Hyper-parameter:dataset" (26.1\%), \\
 & 6.5 (10) & gold answer predicted, wrong attribute & "Result:experimental\_settings" (100\%) \\

\midrule

 & 19.5 (30) & gold answer in main text, table header predicted & "Result:training\_data" (33.3\%), "Result:task" (33.3\%), "Result:model" (33.3\%), \\
 False Positive & 11.7 (18) & partial match, but misses important details & "Result:test\_data" (100\%) \\
 & 6.5 (10) & gold answer in table caption, table header predicted & "Result:metric" (100\%) \\
 & 6.5 (10) & complete mismatch & "Result:experimental\_settings" (100\%) \\

\midrule

Propagated Errors & 3.2 (5) & select wrong record template & "Other:type" (100\%) \\

\bottomrule
\end{tabular}
}
\caption{Error analysis of \method{} (w/ GPT-4) for ML tables, inspecting 591 attribute predictions from 100 cell records sampled from 10 tables. For each fine-grained error type, we provide the error percentage, detailed error sources, and the top three affected attributes.}
\label{tab:sota}
\end{table*}

\paragraph{What errors are made by \method{}?}
To understand where \method{} struggles, we conduct error analysis on GPT-4 predictions for ML tables. We sample 10 tables from the test set and 10 records for each table, comparing each attribute with the gold value. In total, we find 154 errors out of 591 attributes. We group errors into one of eight categories, listed in Table \ref{tab:sota}, with examples presented in Appendix \ref{sec:error_examples}. For instance, one type of false positive error is when the gold attribute value is present in the table caption, but the model is distracted by a table header. Table \ref{tab:sota} provides a detailed breakdown and includes the top three affected attributes within each error category. We find that the most common error occurs when the model fails to identify attributes present in the table (31.2\%), particularly for {\em experimental settings} like {\em 5-shot} in Result records. Another major error is when attributes present in the accompanying text lead to either null predictions (14.9\%) or incorrectly predicting a table header (19.5\%). These errors highlight the challenges of Schema-Driven IE, where the model must understand nuances of table layouts and also effectively integrate information from surrounding text.



\paragraph{How does the data format impact \method{}'s performance?} The variation in model performance across datasets from different domains with unique formats raises questions about the influence of format differences. To address this, we conducted experiments converting ML tables from \LaTeX~to HTML and chemistry tables from XML to CSV, utilizing both commercial (\texttt{tableconvert}\footnote{\url{https://tableconvert.com/api/}}) and open-source (\texttt{TeX4ht}\footnote{\url{https://tug.org/tex4ht/}}) tools, and selecting the one with the highest conversion accuracy. Despite \texttt{tableconvert} showing superior conversion quality, residual code from the original formats in the converted tables, e.g., \LaTeX~commands in HTML tables, presents a novel "code-switching" challenge for \method{}. Performance evaluation with GPT-4 reveals a minimal drop for ML tables (from 74.2 to 74.1 in Table-F\textsubscript{1}) and a more significant decrease for chemistry tables (from 83.4 to 78.1 in Table-F\textsubscript{1}). Both conversion noise and the model's format-specific processing capabilities could contribute to these differences. The optimal performance on original formats underlines the necessity of developing models adept at handling diverse data formats directly, rather than relying on format conversion tools.

\paragraph{Human Evaluation} Similar to our error analysis on ML tables, we manually inspect attribute predictions for 100 cell records for chemistry tables and DisCoMat, as well as 160 pages for SWDE, and report the prediction precision. Half of the sampled data are double-annotated, with the inter-annotator agreement score calculated as the F1 score between the two annotations. The statistics and results are provided in Table \ref{tab:human_eval} in the appendix. The results show that \method{} achieves high precision across different datasets, ranging from 73.9 to 96.4, aligning with the performance under automatic metrics. Additionally, the high inter-annotator agreement scores (all above 90) indicate that the human evaluation is reliable and consistent.

\subsection{Knowledge Distillation}
\label{sec:distill}
Considering the strong performance of API-based models on Schema-Driven IE, we now show that it is possible to use knowledge distillation \citep{le-etal-2022-shot, Kang2023DistillOA} to build a cost-efficient compact model, using ML tables as a demonstration. Specifically, this process first generates synthetic data by performing inference on unlabeled tables using \texttt{code-davinci-002}, followed by fine-tuning a smaller model (e.g., 7B parameters) using the synthetic data.  We compile a collection of 979 arXiv ML papers, submitted between 2008 and 2019, yielding 3,434 tables (containing a total of 100K cells). In Table \ref{tab:distillation}, we can see that LLaMA-7B and Alpaca-7B demonstrate similar performance as seen in the fine-tuning results (Table \ref{tab:test}).  While fine-tuning LLaMA with LoRA \citep{hu2022lora} presents noticeable computational efficiency, full-parameter fine-tuning of T5-11B matches the teacher model's performance.\footnote{
The improvement over the teacher model is not significant (p-value is 42.3\%, \citealp{berg-kirkpatrick-etal-2012-empirical}).
}

\begin{table}[!t]
\centering
\resizebox{1.0\linewidth}{!}{
\begin{tabular}{clcccccc}
\toprule
\multirow{2}{*}{} & \multirow{2}{*}{\textbf{Model (\texttt{GPU hours})}} & \multicolumn{3}{c}{\textbf{Token-Level F\textsubscript{1}}} & \multicolumn{3}{c}{\textbf{EM}} \\
  \cmidrule(l){3-5} \cmidrule(l){6-8} 
& & \textbf{P} & \textbf{R} & \textbf{F\textsubscript{1}} & \textbf{P} & \textbf{R} & \textbf{F\textsubscript{1}} \\ 
\midrule

\multirow{1}{*}{\makecell{Teacher}}
& \texttt{code-davinci-002} & 74.1 & 71.8 & 72.3 & 59.4 & 56.9 & 57.6 \\

\midrule

\multirow{3}{*}{\makecell{Student}}
& LLaMA-7B (\texttt{50h}) & 74.1 & 67.6 & 69.1 & 56.8 & 53.4 & 54.3 \\
& Alpaca-7B (\texttt{50h}) & 72.7 & 64.8 & 67.5 & 56.1 & 50.0 & 52.0 \\
& T5-11B (\texttt{380h}) & 75.8 & 71.4 & 73.2 & 60.3 & 56.7 & 58.1 \\ 

\bottomrule
\end{tabular}
}
\caption{
    Experimental results for knowledge distillation on the ML tables. Student models are trained on 3,434 tables labeled by the teacher model. \texttt{GPU hours} refers to the training time ($\times$ number of GPUs) of student models for one epoch. 
}
\label{tab:distillation}
\end{table}

\subsection{Leaderboards and Image Extraction}

To further validate \textsc{InstrucTE}'s practicality, we integrate it with multi-modal models, like GPT4-V, for extracting data from table images. In an initial study with ML tables, it yields a Table-F\textsubscript{1} of 70.2, approaching the 74.2 Table-F\textsubscript{1} achieved with the original text inputs. Additionally, we explore \textsc{InstrucTE}'s application to the task of Leaderboard Extraction, where it shows competitive performance against leading supervised systems. Due to space constraints, details on these explorations are provided in Appendix \ref{sec:pos_extern}.

\section{Related Work}
\label{sec:related}

\paragraph{Table Understanding in NLP Research} Recently there have been many research efforts involving tables, particularly, table-to-text generation \citep{parikh2020totto, wang-etal-2022-robust, hu-etal-2023-improving}. For example, ToTTo \cite{parikh2020totto} introduced the task of open-domain table-to-text generation. 
In contrast, our work transforms tables into structured JSON records, where a data schema is the only supervision provided.

\paragraph{Pre-training on Semi-structured Data} 
TaPas \cite{herzig-etal-2020-tapas} and TaBERT \cite{yin2020tabert} pre-train on linearized tables with a specialized cell index embedding. TABBIE \cite{iida2021tabbie} employs dual transformers for separate row and column encoding. 
Similarly, TabLLM \citep{pmlr-v206-hegselmann23a} uses general-purpose LLMs to process tables, but we focus on schema-driven IE rather than table classification or question answering.


\paragraph{IE from Semi-structured Data}
Information extraction from semi-structured data has gained increasing interest \citep{33333, dong-etal-2020-multi-modal, gupta2022discomat, lou2023s2abel}. 
OpenCeres \citep{lockard2019openceres} and ZeroShotCeres \citep{lockard-etal-2020-zeroshotceres} highlight open-domain extraction from web data, while AxCell \citep{kardas-etal-2020-axcell} and TDMS-IE \citep{hou-etal-2019-identification} focus on leaderboard extraction from ML tables. DisCoMat \citep{gupta2022discomat} showcases material composition extraction from scientific tables. 
Unlike most existing methods requiring supervised datasets for fine-tuning, our approach stands out by using LLMs to accurately extract data across various domains using an extraction schema. 

\section{Conclusion}
\label{sec:conclu}

This paper explores the capabilities of LLMs for extracting structured data from heterogeneous tables. 
We introduce a new task, Schema-Driven Information Extraction, which converts tables into structured records guided by a human-authored data schema. To facilitate this task, we present a benchmark, comprised of tables from four diverse domains, and evaluate various LLMs through our proposed method \method{}. The experiments reveal that while API-based models excel across domains and formats, open-source models display significant potential in specific areas. Moreover, we conduct detailed ablation studies and analyses to investigate the factors for model success, and validate the feasibility of building compact models through distillation to reduce dependency on APIs. 

\section*{Limitations}

While \method{} showcases strong performance as an instruction-based prompting approach, it encounters specific challenges. Firstly, similar to other prompting methods, its performance could be sensitive to the phrasing of the prompt. 
Despite offering guidelines for crafting prompts in Appendix \ref{sec:promp_appendix}, such as emphasizing clear attribute names, developing robust extraction schemas for new domains often relies on iterative experimentation. Future work could explore automatic prompt optimization \citep{zhou2023large, wang2023promptagent} to reduce the need for human trial-and-error.
Additionally, the model's varying performance across different domains and formats is difficult to interpret, possibly due to biases in the pretraining corpus, a factor we cannot fully analyze due to the opaque nature of the pre-training process. InstrucTE also faces difficulties with dataset-specific nuances, as it operates on general task descriptions without detailed examples, making it challenging to navigate boundary cases effectively. 

Beyond the model's inherent limitations, the availability of specific API-based backbones like GPT-4 and \texttt{code-davinci-002} may change, impacting reliance on these resources. To reduce this dependency, we include results from open-source models and investigate knowledge distillation as a viable alternative, showing promising results. Our benchmark aims to facilitate future research focused on enhancing smaller, openly accessible models, recognizing the importance of such developments for practical application and broader accessibility.

\section*{Ethical Considerations}
Our use of OpenAI's API-based models to distill open-source table extractors complies with OpenAI's terms of service, as we do not {\em ``use the output from the Services to develop models that compete with OpenAI''}. Regarding licenses of four datasets in our \task{} benchmark, the arXiv ML tables align with the licenses of their original papers. The PubMed Chemistry tables, sourced from the PMC Open Access Subset, conform to Creative Commons or equivalent licenses. For the other two datasets, we adapt pre-existing datasets released by the NLP research community, abiding by their respective original licenses.

 \section*{Acknowledgments}
We would like to thank Azure’s Accelerate Foundation Models Research Program and OpenAI’s Researcher Access Program for graciously providing access to API-based models, such as GPT-4. This research is supported in part by the NSF (IIS-2052498), ODNI and IARPA via the HIATUS program (2022-22072200004), and the Defense Advanced Research Projects Agency (DARPA) under Contract No. HR001119C0108. The views, opinions, and/or findings expressed are those of the author(s) and should not be interpreted as representing the official views or policies of the Department of Defense or the U.S. Government. This work is approved for Public Release, Distribution Unlimited.

\bibliography{anthology,custom}
\bibliographystyle{acl_natbib}

\appendix
\clearpage

\section{\method{}}
\label{sec:promp_appendix}

\paragraph{Prompt Formulation}
Our proposed prompt consists of four components: 1) ``Input Table (w/ supp. text)'' includes the table source code paired with supplementary text from the document;
2) ``Extraction Schema'' defines the JSON formats for extracted records, encompassing the record type, attribute names, and associated data types;
3) ``Task-specific Instructions'' outline the task execution process, addressing both the extraction process from individual cells and the traversal strategy across cells, such as \textit{``left-right, top-down''}; 
4) ``Initial Record'' is used to jump-start the prompting process, including the partial record of the first cell.

For ``Input Table (w/ supp. text)'', we employ the BM25 algorithm to retrieve the most relevant paragraphs for each table. For ``Extraction Schema'', we propose two guidelines for schema design: 1) Attribute names should be specific, which decreases the probability of the model generating incorrect attributes, or hallucinations. For instance, when extracting relevant attributes about a movie from a movie webpage, it's advisable to use specific terms such as \texttt{``movie name''} or \texttt{``director name''}, rather than the generic \texttt{``name''}; 2) Attributes should be strategically ordered, placing simpler attributes ahead of more complex ones as errors in preceding attributes can adversely affect the prediction of subsequent ones due to the autoaggressive nature of LMs.
The exact \method{} prompts used in our experiments are shown in Table \ref{tab:full_prompt_ml_chem} and Table \ref{tab:full_prompt_mat_swde}.

\paragraph{Cell Detector} We develop a rule-based method to identify numeric cells for both the ML and chemistry tables. Specifically, for the ML tables, we use the row separator ``\textbackslash\textbackslash'' and the column separator ``\&'' to divide the table into cells. We then loop over each cell, checking for numeric values after stripping away any stylized text. In cases, where a cell contains multiple numeric values, such as ``$0 \pm 0$'', we consistently choose the first numeric value. For the chemistry tables, the parsing process is more straightforward, owing to the structured XML format of the table. Here, we iterate over each cell, verifying if it contains a numeric value once stylized text has been removed. The performance of our rule-based cell detector on two datasets is presented in Table \ref{tab:cell_detect}. In the case of \discomat{}, we use the cell detector provided by the original paper \citet{gupta2022discomat}.

\begin{table}[h!]
\small
\begin{center}
\begin{tabular}{lcccc}
\toprule
\textbf{Dataset} & \textbf{Split} & \textbf{P} & \textbf{R} & \textbf{F\textsubscript{1}} \\
\midrule

\multirow{2}{*}{ML Tables} & Dev & 100.0 & 97.0 & 98.0 \\
 & Test & 99.9 & 99.6 & 99.7 \\

\midrule

\multirow{2}{*}{Chem. Tables} & Dev & 100.0 & 100.0 & 100.0 \\
 & Test & 100.0 & 98.3 & 99.2 \\

\bottomrule
\end{tabular}
\end{center}
\caption{\label{tab:cell_detect} Results of (numeric) cell detection on ML and chemistry tables.
}
\end{table}

\begin{table*}[th!]
\small
\centering
\scalebox{1}{
\begin{tabular}{lL{13cm}}
\toprule 

 \textbf{Dataset} & \textbf{Full Prompt}  \\
 
\midrule

ML Tables & 





\includegraphics[width=0.8\textwidth]{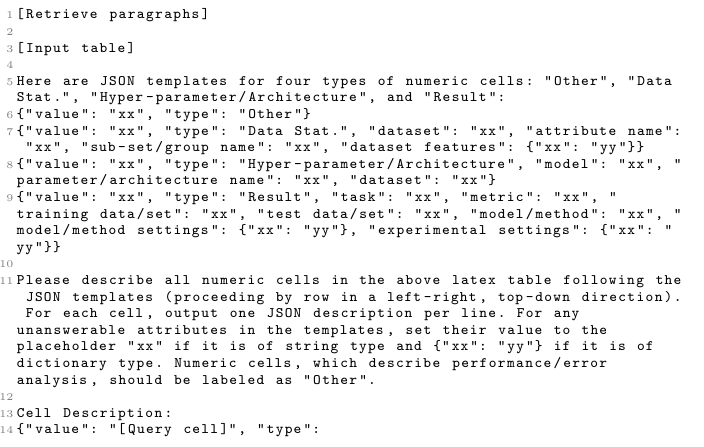}

\\

\midrule

Chem. Tables & 





\includegraphics[width=0.8\textwidth]{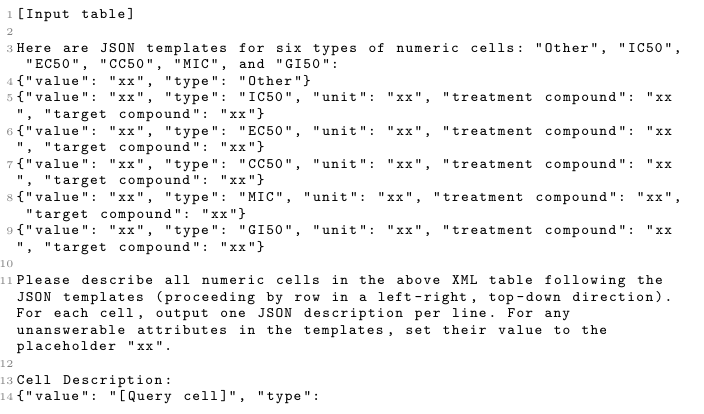}

\\ 

 \bottomrule
\end{tabular}
}

\caption{\label{tab:full_prompt_ml_chem} \method{} prompts used for ML and chemistry tables.
}
\end{table*}

\begin{table*}[th!]
\small
\centering
\scalebox{1}{
\begin{tabular}{lL{13cm}}
\toprule 

 \textbf{Dataset} & \textbf{Full Prompt}  \\
 
\midrule

\discomat{} & 





\includegraphics[width=0.8\textwidth]{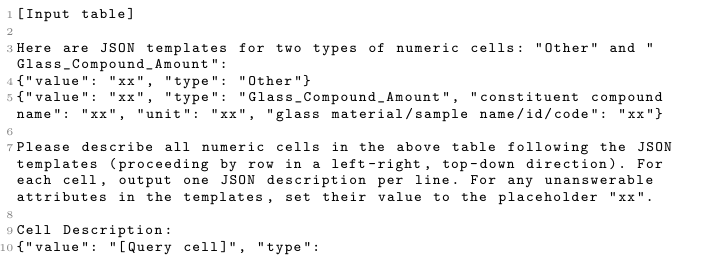}

\\

\midrule

SWDE-auto & 




\includegraphics[width=0.8\textwidth]{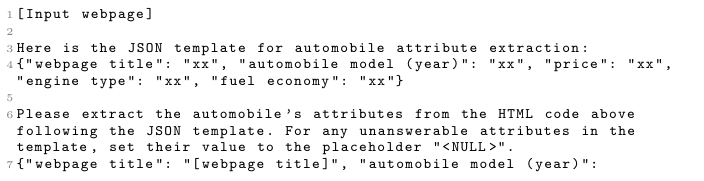}

\\ 

 \bottomrule
\end{tabular}
}

\caption{\label{tab:full_prompt_mat_swde} \method{} prompts used for \discomat{} and SWDE. For SWDE, we use the ``Auto'' vertical as an illustrative example, and the prompts for other verticals differ only in attribute names (refer to Table \ref{tab:swde_stats} for the attributes of each vertical).
}
\end{table*}

\begin{table}[ht]
\centering
\scalebox{0.6}{
\begin{tabular}{cccc}
\toprule 
\textbf{Vertical} & \textbf{\# Sites} & \textbf{\# Pages} & \textbf{Attributes} \\ 
\midrule
Auto & 10 & 17,923 & \texttt{model, price, engine, fuel-economy} \\ \hline
Book & 10 & 20,000 & \texttt{\makecell{title, author, ISBN-13,\\ publisher, publish-date}} \\ \hline
Camera & 10 & 5,258 & \texttt{model, price, manufacturer} \\ \hline
Job & 10 & 20,000 & \texttt{title, company, location, date} \\ \hline
Movie & 10 & 20,000 & \texttt{title, director, genre, rating} \\ \hline
NBA Player & 10 & 4,405 & \texttt{name, team, height, weight} \\ \hline
Restaurant & 10 & 20,000 & \texttt{name, address, phone, cuisine} \\ \hline
University & 10 & 16,705 & \texttt{name, phone, website, type} \\
\bottomrule
\end{tabular}
}
\caption{SWDE statistics. } 
\label{tab:swde_stats}
\end{table}






\section{arXiv Machine Learning Tables}
\label{sec:mltab_appendix}

\paragraph{Extraction Attributes} We design a set of extraction attributes for each of the three primary types of numeric cells in ML tables: ``Result'', ``Hyper-parameter'', and ``Data Statistics''. These attributes are outlined in detail below.

\begin{itemize}[parsep=3pt,leftmargin=0.5cm]
    \item ``Result'' includes seven attributes: \texttt{training data}, \texttt{test data}, \texttt{task}, \texttt{metric}, \texttt{model}, \texttt{model settings} and \texttt{experimental settings}. The first five attributes are fixed, with answers being text spans in the paper. The last two attributes, \texttt{model settings} and \texttt{experimental settings}, are free-form attributes, with answers being JSON objects. 
    For example, the \texttt{experimental settings} attribute may be \textit{\{``number of training examples'': ``0''}\} for a zero-shot setting.
    This scheme is more detailed than previous approaches \citep{hou-etal-2019-identification, kardas-etal-2020-axcell} and can accommodate a broader range of ML paradigms and provide more granular information.

    \item ``Hyper-parameter'' includes optimization parameters like learning rate and batch size, as well as numeric descriptions of model architectures such as layer count. The three fixed attributes for this category are: \texttt{model}, \texttt{parameter/architecture}, and \texttt{dataset}.

    \item ``Data Stat.'' covers four attributes: \texttt{dataset}, \texttt{dataset attribute}, \texttt{sub-set/group}, and \texttt{dataset features}. The \texttt{sub-set/group} specifies a dataset subset (e.g., ``train'' or ``test''), while \texttt{dataset features}, a free-form attribute, captures various dataset characteristics like the language or domain.

\end{itemize}

\paragraph{Annotation Process} We sample 10 papers from each of three pertinent arXiv fields: Machine Learning, Computer Vision, and Natural Language Processing. After removing papers without \LaTeX{} source code or any tables, a total of 25 papers are covered in our dataset. To optimize the annotation budget and the dataset diversity, we cap the number of annotated tables to five per paper. Recognizing the domain-specific expertise needed, we employ expert annotators with backgrounds in ML research, who are provided with tables in both \LaTeX~and PDF formats and encouraged to thoroughly read the paper before annotation. The annotation process comprises two steps: 1) identifying the numeric cells and their record types, and 2) filling in the slots of pre-determined attributes, forming a JSON record with keys as attribute names and values as extracted content, in a text editor. Consequently, the dataset contains 122 tables, with 3,792 cells and 21K attributes annotated. 




\begin{table}[h!]
\small
\begin{center}
\scalebox{0.9}{
\begin{tabular}{lcccc}
\toprule
\textbf{Dataset} & \textbf{\# Records} & \textbf{\# Attr.} & \textbf{Precision} & \textbf{IAA} \\
\midrule

ML Tables & 100 & 591 & 73.9 & 95.7 \\
Chem. Tables & 100 & 380 & 95.3 & 100\\
\textsc{DisCoMat} & 100 & 201 & 92.5 & 99.4 \\
SWDE & 160 & 640 & 96.4 & 98.2 \\

\bottomrule
\end{tabular}
}
\end{center}
\caption{\label{tab:human_eval} Statistics and results of attribute-level human evaluation on four datasets. The inter-annotator agreement score (IAA) is calculated as the F1 score between the two annotations
}
\end{table}

\begin{figure}[t]
    \centering
    \includegraphics[scale=0.53]{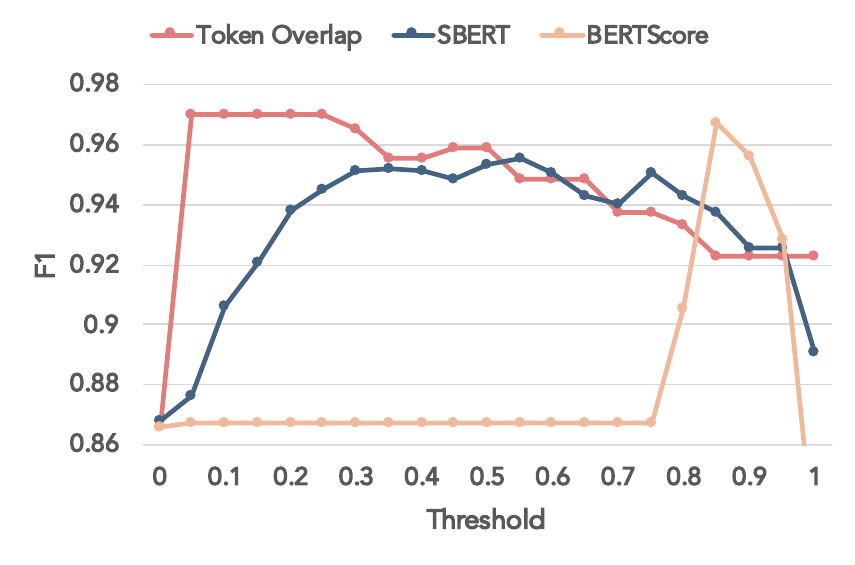}
    \caption{
    Results of comparing various metrics, including token-level F\textsubscript{1}, SBERT, and BERTScore, to human judgment over different thresholds on ML tables. Numbers are computed over 677 sampled attributes that are paired with respective gold references.
    }
    \label{fig:metrics}
\end{figure}

\begin{figure*}[!ht]
    \centering
    \includegraphics[width=\textwidth]{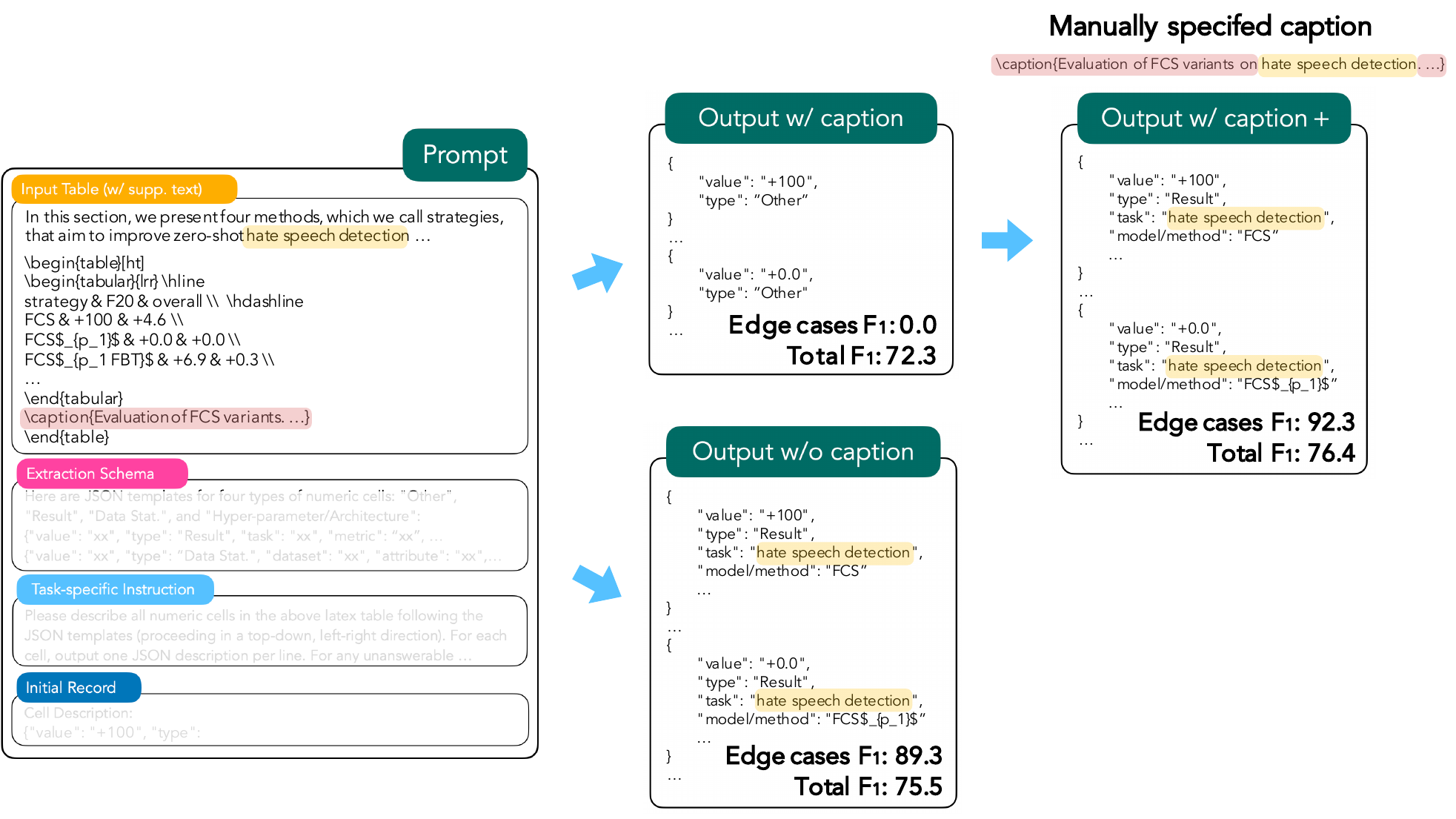}
    \caption{
     An error analysis of edge cases in which the predictions made by \textsc{InstrucTE} with captions default to ``Other'' (resulting in an 0 F\textsubscript{1}). Our hypothesis that this issue may stem from the caption's lack of specificity is tested by manually expanding the caption (displayed on the right). This amendment significantly improves the performance on these edge cases, increasing the F\textsubscript{1} score to 92.3.
    }
    \label{fig:analysis_of_caption}
\end{figure*}

\begin{figure*}[ht!]
    \centering
    \includegraphics[width=\textwidth]{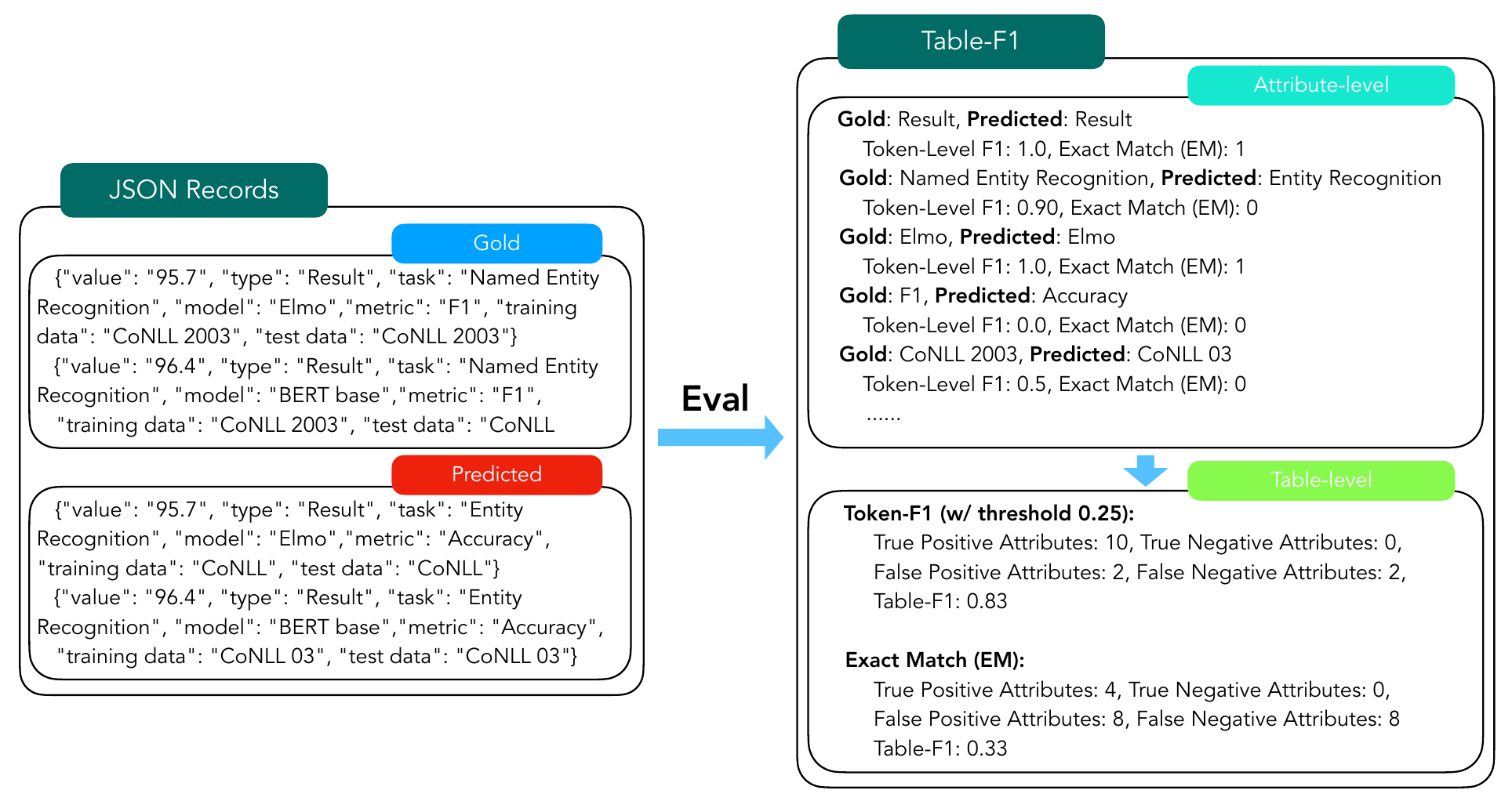}
    \caption{An example of Table-F\textsubscript{1} calculation, where two predicted records are compared against the two gold records.
    }
    \label{fig:metric}
\end{figure*}

\section{Evaluation Metrics}
\label{sec:eval_metric_app}

Comparing an LLM-predicted JSON object with a gold JSON object is a non-trivial task, as those generative LLMs may produce text spans that do not exactly exist in the input table. Consequently, we devote substantial effort to examining various metrics to determine the one best suited for our task using ML tables. Here, we consider three metrics: the standard token-level F\textsubscript{1} to capture the level of lexical overlap between the predicted and gold attributes, and two semantic similarity metrics, SBERT \citep{reimers-2019-sentence-bert} and BERTScore \citep{bert-score}, to identify semantically similar expressions (e.g., \# params vs. the number of parameters). 

\paragraph{Meta Evaluation}
To assess how accurate each metric is compared to human evaluation, we manually annotated predicted-gold attribute pairs as to whether or not each pair matches. We consider a given pair to ``match'' if they are semantically equivalent, meaning they can be used interchangeably. For attributes that encapsulated multiple sub-attributes, we consider a pair to match if at least half of the sub-attributes are matched (i.e., F\textsubscript{1} score $\geq$ 0.5), with the decision for each sub-attribute being based on the same as in the text-span attributes. For the set of pairs to annotate and use as a test set, we sample a total of 100 cell pairs (i.e., 677 attribute pairs) according to the following process: 1) we first uniformly sample a table from the development set (containing 10 papers); and 2) we then sample a random cell from the table, ensuring there were no duplicate cells. For each pair of predicted-gold attributes, each metric's decision (1 or 0) is made using a specific threshold. For example, if the token-level F\textsubscript{1}'s score for paired attributes is 0.4 and the threshold is 0.5, then the decision would be 0, indicating no match. The decisions over the test set containing 677 attribute pairs are then compared to human evaluation. In this binary classification problem, F\textsubscript{1} is used to evaluate the performance of the metrics.

In Table \ref{tab:metrics}, we present the performances of each metric with the optimal threshold for each. Surprisingly, we find that the token-level F\textsubscript{1} (with a threshold of 0.25) decision aligns nearly perfectly with human judgment, and performs the best among all metrics for our task. This might suggest that discerning subtle differences is more crucial than identifying different phrases with the same meaning for this task. Based on these empirical findings, we opt for the token-level F\textsubscript{1} for automatic evaluation at the attribute level. This choice is highly desirable not only because of its high accuracy but also due to its simplicity.

\begin{table}[t]
\centering
\scalebox{0.75}{
\begin{tabular}{lccc}
\toprule 
 & \textbf{token-level F\textsubscript{1}} & \textbf{SBERT} & \textbf{BERTScore} \\ 
\midrule
Meta Eval. F\textsubscript{1} & 97.0 & 95.6 & 96.7 \\
Threshold & 0.25 & 0.55 & 0.85 \\
\bottomrule
\end{tabular}
}
\caption{ 
Results of comparing various metrics, including token-level F\textsubscript{1}, SBERT, and BERTScore, to human judgment on ML tables. 
Numbers are computed over 677 sampled attributes that are paired with gold references. The highest achieved F\textsubscript{1} scores are displayed alongside the thresholds. A complete illustration of results, sorted by thresholds, can be found in Figure \ref{fig:metrics} in Appendix.
}
\label{tab:metrics}
\end{table}

\section{Implementation Details}
\label{sec:implement_details}

Considering the lengthy source code for tables, we employ different strategies to encode the input table and perform Schema-Driven IE, based on the context length of the chosen LLM. For LLMs with a larger context length, such as GPT-4, \texttt{code-davinci-002}, and CodeLlama, we input the full table and conduct the proposed error recovery process. For LLMs with a more limited context length, such as LLaMA and T5-11B, we query each target cell individually. The input table is condensed by rows, retaining the first two rows, typically containing headers, and the row with the query cell, with the token \texttt{<select>} pinpointing the position of the query cell. We use greedy decoding to maximize the reproducibility of our results.

For the TableQA setting, we divide the problem into two steps: selecting the record type and predicting the relevant attributes. For T5 and Flan-T5, the first step is modeled as a multi-choice QA problem, where the model chooses the type of the query cell from a list of provided options. The second step is modeled as an extractive QA task, asking the model to pinpoint the answer spans for the attributes associated with the selected type. For TaPas, the initial step is treated as a classification problem, whereas the latter one is handled as a cell selection problem. The hyper-parameters used for fine-tuning T5 and TaPas are presented in Table \ref{tab:hyperparam_ft}.

\begin{table}[t]
\centering
\scalebox{0.75}{
\begin{tabular}{lcc}
\toprule 
 & \textbf{T5 (11B)} & \textbf{TaPas}  \\ 
\midrule
learning rate & 1e-4 & 5e-5  \\
batch size & 8 & 32  \\
\# epoches  & 5 & 10 \\
\bottomrule
\end{tabular}
}
\caption{Hyper-parameters used for fine-tuning T5 and TaPas.
}
\label{tab:hyperparam_ft}
\end{table}

\section{Error Analysis of Caption}
\label{sec:error_analysis_of_caption}

In Section \ref{sec:ablation}, we observe an unexpected finding that table captions do not enhance performance, but rather seem to detract from it, which is counterintuitive. To delve deeper into this observation, we conduct an error analysis. This involves comparing the performances of our \textsc{InstrucTE} system with and without captions at the table level. This analysis uncovers a few outliers (3 out of 68) where including a caption leads to a 0 F\textsubscript{1} score, whereas the score is near perfect when the caption is excluded.
For instance, as depicted in Figure \ref{fig:analysis_of_caption}, the predictions all fall into the ``Other'' category when a caption is included, leading to a 0 F\textsubscript{1} score in these outlier instances. Conversely, removing the caption results in an F\textsubscript{1} score of 89.3. This high score is due to the fact that retrieved paragraphs provide ample contextual information (e.g., ``hate speech detection'') without the presence of a caption.

We hypothesize that the model's inclination to predict ``Other'' in the presence of a caption may be a consequence of the captions' lack of specificity with respect to the attributes relevant to the table cells (for example, ``hate speech detection''). This lack of explicit, relevant details could create confusion in associating the caption with the retrieved paragraphs, thereby misleading the model.
To test our hypothesis, we manually adjust the captions to include more specific attributes, such as ``hate speech detection'' and ``T5-Base.'' As a result, we observe an improvement in the model's performance with the revised caption, with the total F\textsubscript{1} score even exceeding that achieved without a caption. This outcome partially supports our hypothesis and suggests that carefully crafted captions could indeed be beneficial, aligning with our initial expectations. However, this investigation also points to the fact that the model currently lacks robustness in handling these outlier scenarios.

\section{\method{} Errors}
\label{sec:error_examples}

This section presents examples of \method{}'s errors (w/ GPT-4) on ML tables, illustrating each of the eight fine-grained error types.

\subsection{False Negative: gold answer in table, not predicted}

\noindent\textbf{Input} (Table 5 of arXiv paper 2210.00044v1)\vspace{-70pt}

\hspace{-1cm}\includegraphics[width=0.5\textwidth]{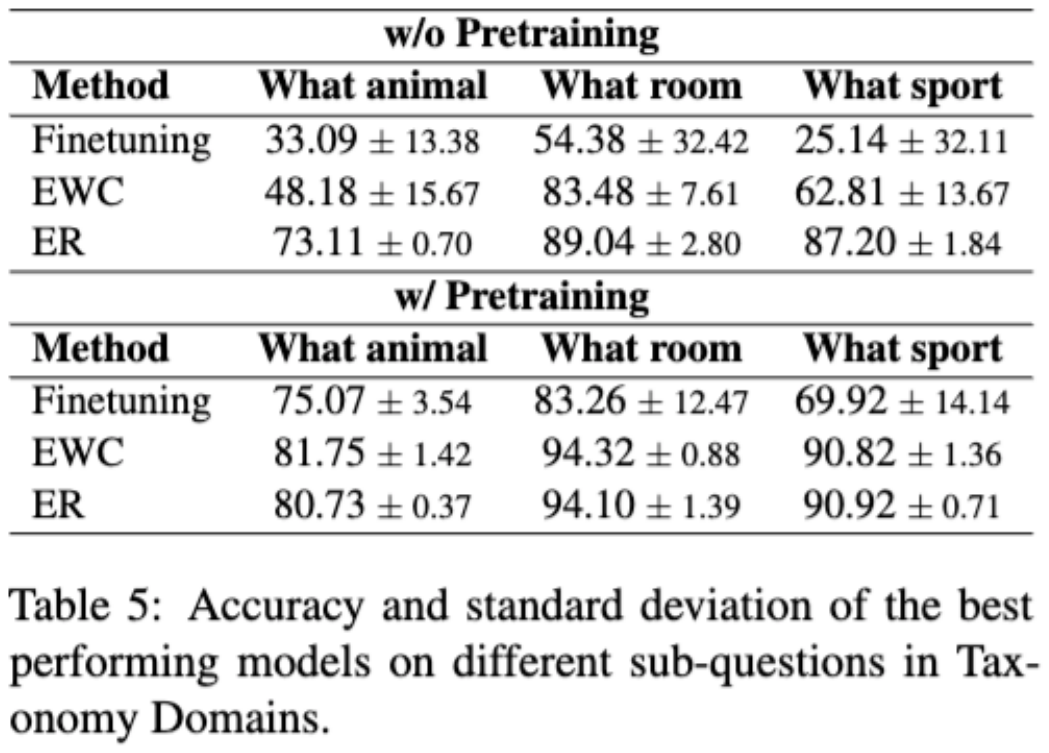}\vspace{-70pt}

\noindent\textbf{Predicted Records}
\lstset{xleftmargin=1em}
\begin{lstlisting}[]
{"value": "83.48", ..., "model/method settings": {"xx": "yy"}, ...}
{"value": "62.81", ..., "model/method settings": {"xx": "yy"}, ...}
    ...
{"value": "94.32", ..., "model/method settings": {"xx": "yy"}, ...}
    ...
\end{lstlisting}

\noindent\textbf{Gold Records}

\begin{lstlisting}[]
{"value": "83.48", ..., "model/method settings": {"w/o Pretraining": "true"}, ...}
{"value": "62.81", ..., "model/method settings":{"w/o Pretraining": "true"}, ...}
    ...
{"value": "94.32", ..., "model/method settings": {"w Pretraining": "true"}, ...}
    ...
\end{lstlisting}

\subsection{False Negative: gold answer in main text, not predicted}

\noindent\textbf{Input} (Table 4 of arXiv paper 2210.00193v1)\vspace{-90pt}

\hspace{-1cm}\includegraphics[width=0.5\textwidth]{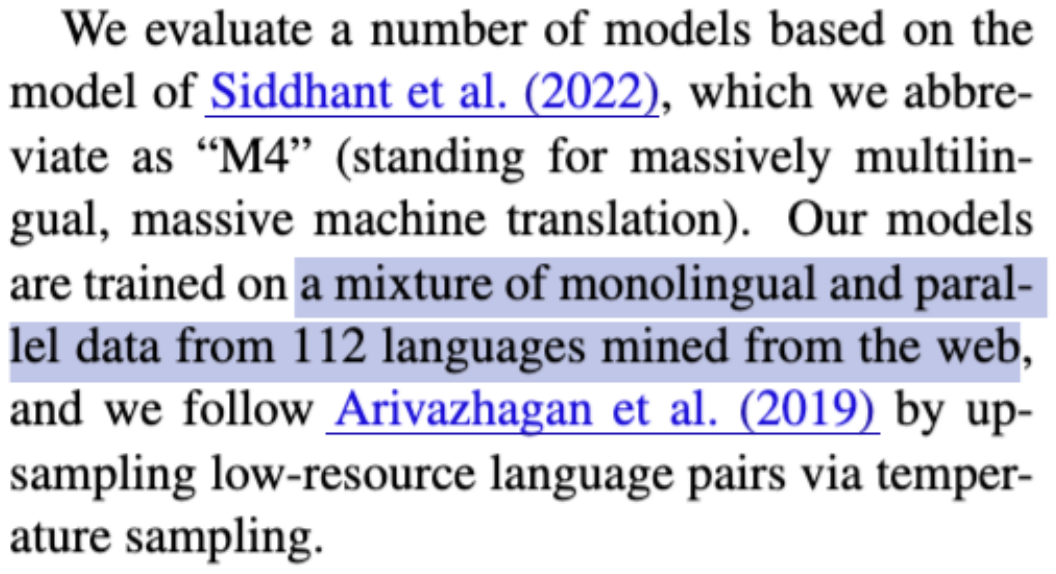}\vspace{-160pt}
\hspace{-3cm}\includegraphics[width=0.5\textwidth]{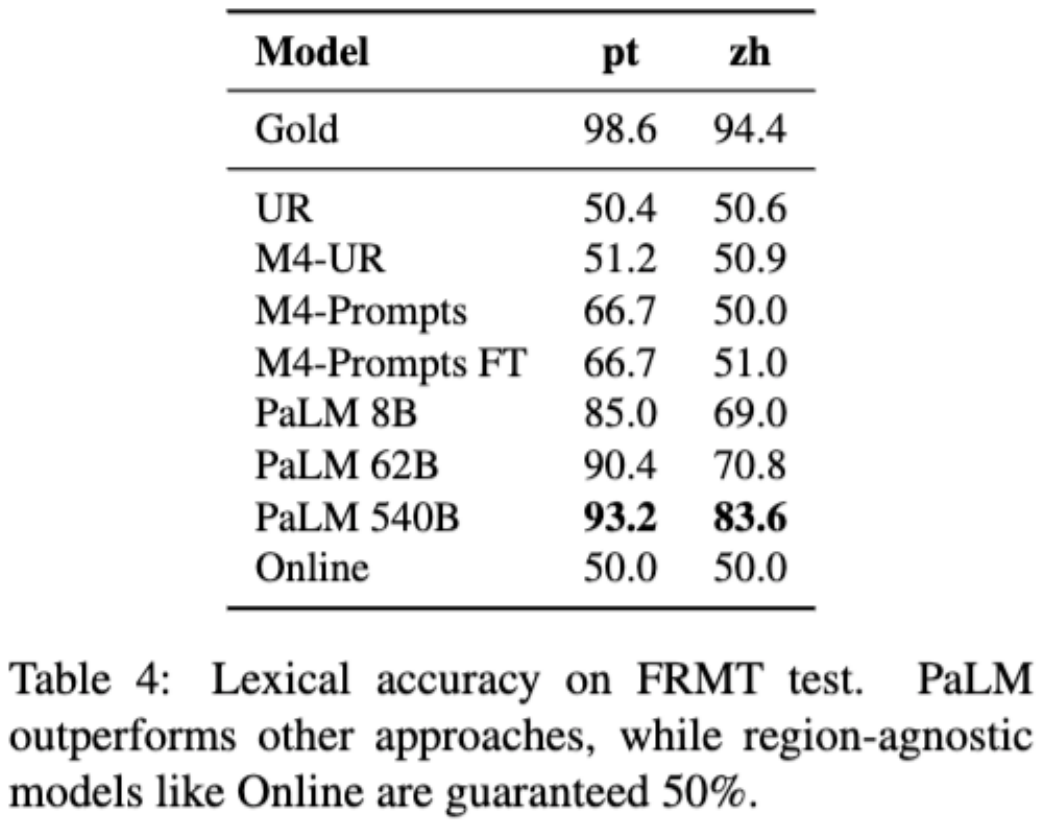}\vspace{-70pt}

\noindent\textbf{Predicted Records}

\begin{lstlisting}[]
{"value": "51.2", ..., "training data/set": "xx", ...}
{"value": "50.9", ..., "training data/set": "xx", ...}
    ...
{"value": "66.7", ..., "training data/set": "xx", ...}
    ...
\end{lstlisting}

\noindent\textbf{Gold Records}

\begin{lstlisting}[]
{"value": "51.2", ..., "training data/set": "a mixture of monolingual and parallel data from 112 languages mined from the web", ...}
{"value": "50.9", ..., "training data/set": "a mixture of monolingual and parallel data from 112 languages mined from the web", ...}
    ...
{"value": "66.7", ..., "training data/set": "a mixture of monolingual and parallel data from 112 languages mined from the web", ...}
    ...
\end{lstlisting}

\vspace{30pt}
\subsection{False Negative: gold answer predicted, wrong attribute}

\noindent\textbf{Input} (Table 5 of arXiv paper 2210.00044v1)\vspace{-70pt}

\hspace{-1cm}\includegraphics[width=0.5\textwidth]{figures/error/fn1.pdf}\vspace{-70pt}

\noindent\textbf{Predicted Records}

\begin{lstlisting}[]
{"value": "83.48", ..., "model": "EWC", "experimental settings": {"xx": "yy"}, ...}
{"value": "62.81", ..., "model": "EWC", "experimental settings": {"xx": "yy"}, ...}
    ...
{"value": "94.32", ..., "model": "EWC", "experimental settings": {"xx": "yy"}, ...}
    ...
\end{lstlisting}

\noindent\textbf{Gold Records}

\begin{lstlisting}[]
{"value": "83.48", ..., "experimental settings": {"Method": "EWC"}, ...}
{"value": "62.81", ..., "experimental settings": {"Method": "EWC"}, ...}
    ...
{"value": "94.32", ..., "experimental settings": {"Method": "EWC"}, ...}
    ...
\end{lstlisting}

\subsection{False Positive: gold answer in main text, table header predicted}

\noindent\textbf{Input} (Table 5 of arXiv paper 2210.00044v1)\vspace{-90pt}

\hspace{-1cm}\includegraphics[width=0.5\textwidth]{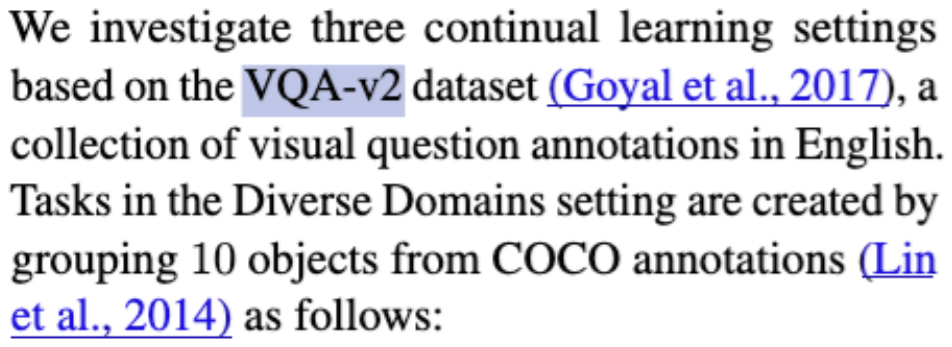}\vspace{-180pt}
\hspace{-2cm}\includegraphics[width=0.45\textwidth]{figures/error/fn1.pdf}\vspace{-70pt}

\noindent\textbf{Predicted Records}

\begin{lstlisting}[]
{"value": "83.48", ..., "training data/set": "What room", ...}
{"value": "62.81", ..., "training data/set": "What sport", ...}
    ...
{"value": "94.32", ..., "training data/set": "What room", ...}
    ...
\end{lstlisting}

\noindent\textbf{Gold Records}

\begin{lstlisting}[]
{"value": "83.48", ..., "training data/set": "VQA-v2", ...}
{"value": "62.81", ..., "training data/set": "VQA-v2", ...}
    ...
{"value": "94.32", ..., "training data/set": "VQA-v2", ...}
    ...
\end{lstlisting}

\subsection{False Positive: partial match, but misses important details}

\noindent\textbf{Input} (Table 4 of arXiv paper 2210.00193v1) \vspace{-30pt}

\hspace{-0.5cm}\includegraphics[width=0.45\textwidth]{figures/error/fn2_table.pdf}\vspace{-50pt}

\noindent\textbf{Predicted Records}

\begin{lstlisting}[]
{"value": "51.2", ..., "test data/set": "FRMT test", ...}
{"value": "50.9", ..., "test data/set": "FRMT test", ...}
    ...
{"value": "66.7", ..., "test data/set": "FRMT test", ...}
    ...
\end{lstlisting}

\noindent\textbf{Gold Records}

\begin{lstlisting}[]
{"value": "51.2", ..., "test data/set": "FRMT pt", ...}
{"value": "50.9", ..., "test data/set": "FRMT zh", ...}
    ...
{"value": "66.7", ..., "test data/set": "FRMT pt", ...}
    ...
\end{lstlisting}

\subsection{False Positive: gold answer in table caption, table header predicted}

\noindent\textbf{Input} (Table 3 of arXiv paper 2210.00740v1)\vspace{-35pt}

\hspace{-0.5cm}\includegraphics[width=0.45\textwidth]{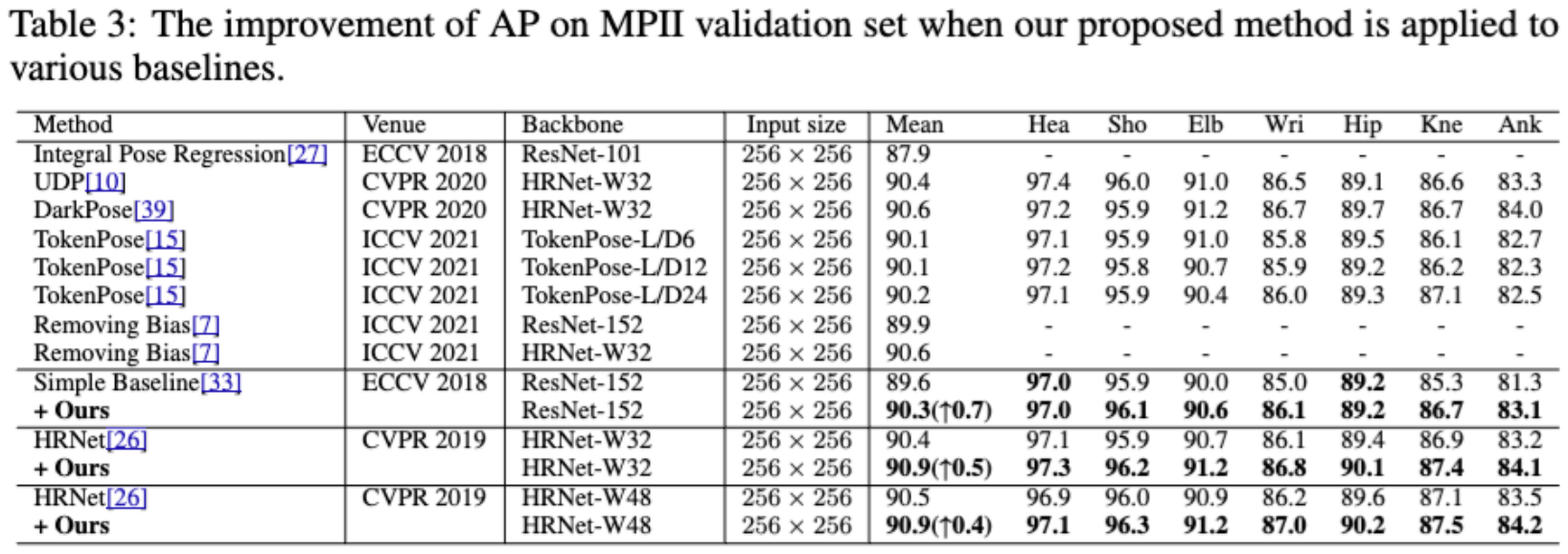}\vspace{-30pt}

\noindent\textbf{Predicted Records}

\begin{lstlisting}[]
{"value": "97.2", ..., "metric": "Hea", ...}
{"value": "86.7", ..., "metric": "Wri", ...}
    ...
{"value": "82.3", ..., "metric: "Ank", ...}
    ...
\end{lstlisting}

\noindent\textbf{Gold Records}

\begin{lstlisting}[]
{"value": "97.2", ..., "metric": "AP", ...}
{"value": "86.7", ..., "metric": "AP", ...}
    ...
{"value": "82.3", ..., "metric": "AP", ...}
    ...
\end{lstlisting}

\subsection{False Positive: complete mismatch}

\noindent\textbf{Input} (Table 3 of arXiv paper 2210.00740v1)\vspace{-35pt}

\hspace{-0.5cm}\includegraphics[width=0.45\textwidth]{figures/error/fp3.pdf}\vspace{-30pt}

\noindent\textbf{Predicted Records}

\begin{lstlisting}[]
{"value": "97.2", ..., "experimental settings": {"Venue": "CVPR 2020"}, ...}
{"value": "86.7", ..., "experimental settings": {"Venue": "CVPR 2020"}, ...}
    ...
{"value": "82.3", ..., "experimental settings": {"Venue": "ICCV 2021"}, ...}
    ...
\end{lstlisting}

\noindent\textbf{Gold Records}

\begin{lstlisting}[]
{"value": "97.2", ..., "experimental settings": {"xx": "yy"}, ...}
{"value": "86.7", ..., "experimental settings": {"xx": "yy"}, ...}
    ...
{"value": "82.3", ..., "experimental settings": {"xx": "yy"}, ...}
    ...
\end{lstlisting}

\subsection{Propagated Errors: select a wrong record template
}

\noindent\textbf{Input} (Table 7 of arXiv paper 2210.00627v1)\vspace{-20pt}

\hspace{-0.5cm}\includegraphics[width=0.45\textwidth]{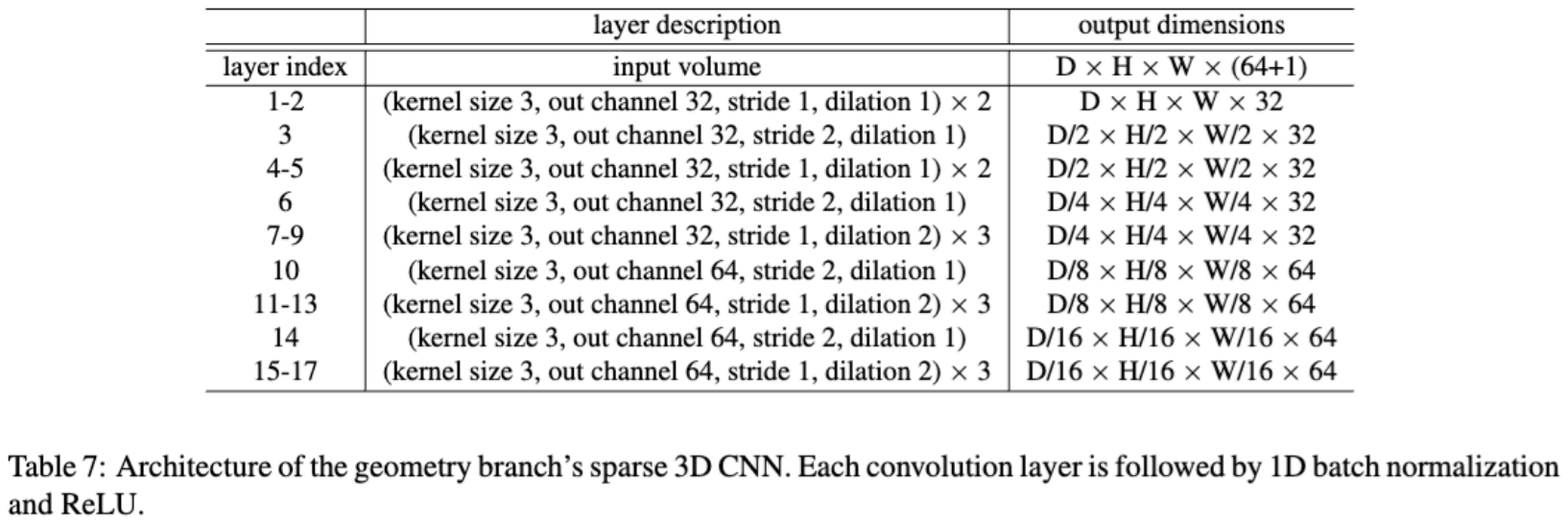}

\noindent\textbf{Predicted Records}

\begin{lstlisting}[]
{"value": "3", "type": "Hyper-parameter", ...}
{"value": "6", "type": "Hyper-parameter", ...}
    ...
{"value": "14", "type": "Hyper-parameter", ...}
\end{lstlisting}

\noindent\textbf{Gold Records}

\begin{lstlisting}[]
{"value": "3", "type": "Other"}
{"value": "6", "type": "Other"}
    ...
{"value": "14", "type": "Other"}
\end{lstlisting}

\section{Extracting Leaderboards from Table Images}
\label{sec:pos_extern}

\subsection{Extraction from Table Images}
One practical challenge with \method{} is the need for tables in a textual format, while many tables are available only as PDFs or images.  
To address this, we integrate \method{} with multi-modal models to extract structured data from table images. Specifically, we experiment with two strategies: 1) direct extraction from table images, and 2) a pipeline that first employs multi-modal models to transform table images into text, and then run \textsc{InstrucTE} on the textual tables.



In a preliminary study with ML tables, we use GPT-4V as the backbone for \method{}. We find that the pipeline method yields a Table-F\textsubscript{1} score of 70.2 from image inputs, approaching the 74.2 Table-F\textsubscript{1} achieved with the original text inputs. It outperforms direct extraction using GPT-4V, which attains only a Table-F\textsubscript{1} score of 46.4, as the pipeline can capitalize on \textsc{InstrucTE}'s error recovery capabilities, resulting in more thorough and accurate extractions. 

Additionally, we test \texttt{IDEFICS-80b-instruct} \citep{laurençon2023obelics}, a leading open-source multi-modal model, which unfortunately could not perform the table-text conversion or direct extraction.\footnote{The \texttt{IDEFICS-80b-instruct} model either produces unrelated content or simply output "I am sorry, but I cannot generate LaTeX code from the table."} This suggests a clear avenue for future research to enhance multi-modal models' ability to accurately process image-based tables.

\begin{figure*}[ht!]
    \centering
    \includegraphics[width=\textwidth]{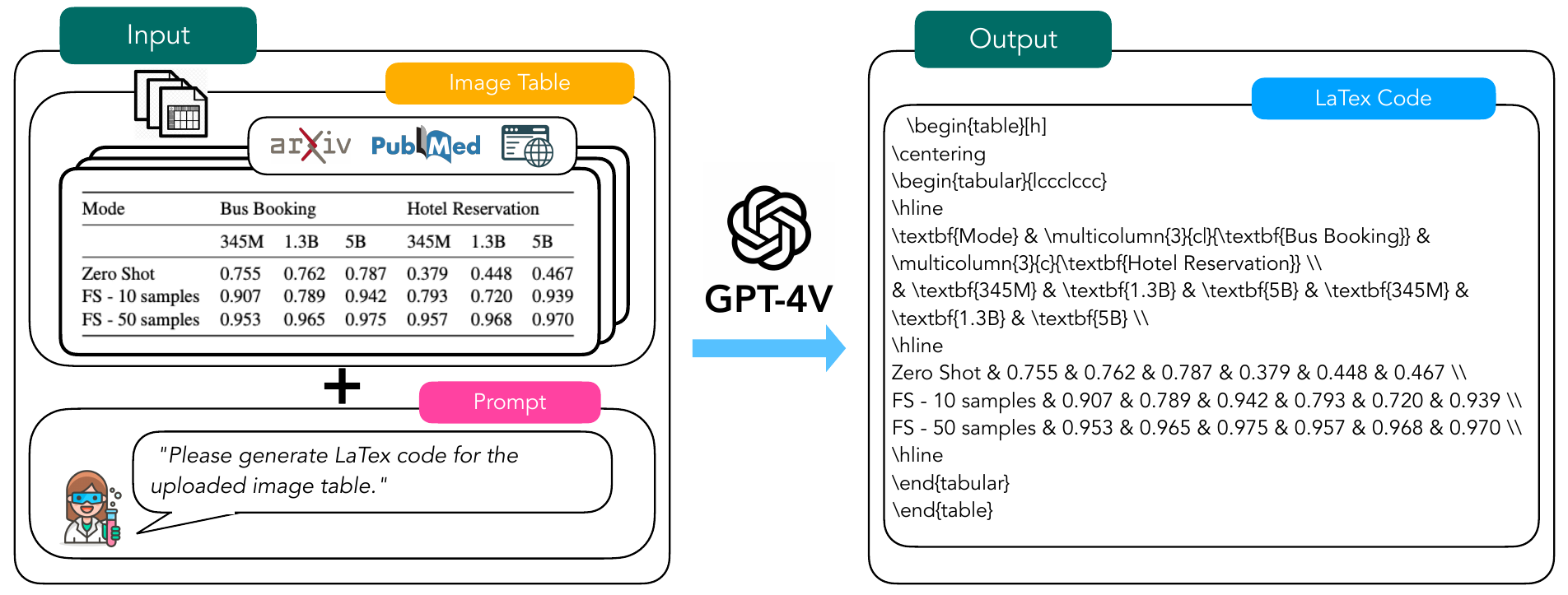}
    \caption{Generate \LaTeX~code for image tables using GPT-4V. 
    }
    \label{fig:gpt-4v}
\end{figure*}

\subsection{Leaderboard Extraction from ML Papers}
\label{sec:axcell_details}

\paragraph{Task Definition \& SOTA Methods}
The task of leaderboard extraction \citep{hou-etal-2019-identification, kardas-etal-2020-axcell} entails extracting leaderboard tuples (\texttt{task}, \texttt{dataset}, \texttt{metric}, \texttt{score}) from tables in ML papers. Unlike our proposed Schema-Driven IE, which requires open-domain span identification, leaderboard extraction presumes prior knowledge of all leaderboards, represented as pre-defined (\texttt{task}, \texttt{dataset}, \texttt{metric}) tuples, and centers on linking numeric cells to these leaderboards. 

The state-of-the-art leaderboard extraction method, \axcell{} \cite{kardas-etal-2020-axcell}, is a comprehensive pipeline system comprising four components: Table Type Classification, Table Segmentation, Cell Linking, and Filtering. For each component, except the last one, \axcell{} employs a supervised model. It starts with table type classification to identify result-related tables, which are then passed to the table segmenter responsible for annotating the header cells of the table. Following this step, a retrieval model links numeric cells in the table to pre-defined leaderboards using human-engineered features. Lastly, \axcell{} filters and selects the best record based on the leaderboard taxonomy criteria, such as retaining higher values for "Accuracy" and lower ones for "error rate".

\paragraph{Application of \method{}}
To extract leaderboards from an ML paper, we consider all tables that contain numeric cells, instead of selecting tables via a trained classifier as in \axcell{}. For each table, we run \method{} using a customized leaderboard extraction JSON template. This template resembles the ML-table template with two additional fixed attributes: \texttt{eval split} and \texttt{eval class} in the ``Result'' cell template. We add the \texttt{eval split} attribute because the evaluated split is essential information for this task; for instance, \textit{``dev F\textsubscript{1}''} and \textit{``test F\textsubscript{1}''} are treated as different metrics in the leaderboard taxonomy. 
The \texttt{eval class} attribute is used to exclude sub-set or sub-class results that are typically present in analysis tables. After generating all predicted cell descriptions, we filter them based on three criteria: 1) the \texttt{type} attribute must be \textit{``Result''}; 2) the \texttt{eval class} attribute must be \textit{``all''} or \textit{``Null''} as observed on the development set; and 3) the cell must be bolded in the table, as this usually indicates its superior performance and possible relevance to the leaderboard. 
For papers without any bolded cells, we experiment with two strategies: 1) include all the remaining cells in the table that meet the first two criteria; 2) use cells selected by \axcell{}, as its engineered features for cell selection may be useful. This hybrid system is referred to as \method{}+.
We then use the predicted \texttt{task}, \texttt{dataset}, and \texttt{metric} attributes in each JSON record to match with the pre-defined leaderboards using token-level F\textsubscript{1}, and we select the leaderboard with the highest average score over three attributes. Finally, following \axcell{}, we choose the best record based on the leaderboard taxonomy criteria, e.g., retaining higher values for "Accuracy" and lower ones for "error rate".


\begin{table}[t]
\centering
\scalebox{0.7}{
\begin{tabular}{lcccccc}
\toprule 
\multirow{2}{*}{\textbf{Method}} & \multicolumn{3}{c}{\textbf{Micro-Average}} & \multicolumn{3}{c}{\textbf{Macro-Average}} \\
\cmidrule(l){2-4} \cmidrule(l){5-7} 
& \textbf{P} & \textbf{R} & \textbf{F\textsubscript{1}} & \textbf{P} & \textbf{R} & \textbf{F\textsubscript{1}} \\ 
\midrule
\axcell{} & \textbf{25.4} & 18.4 & 21.3 & \textbf{21.5} & 21.5 & 20.0 \\
\method{} & 20.1 & 20.8 & 20.5 & 20.3 & 23.1 & 19.6 \\
\method{}+ & 23.9 & \textbf{21.2} & \textbf{22.4} & 21.2 & \textbf{23.7} & \textbf{20.5} \\
\bottomrule
\end{tabular}
}
\caption{Leaderboard extraction results on the \textsc{PWC Leaderboards} dataset. 
}
\label{tab:pwd_results}
\end{table}

\paragraph{Results} We compare \method{} with \axcell{} on \textsc{PWC Leaderboards} \cite{kardas-etal-2020-axcell}, the largest dataset for leaderboard extraction. For \method{}, we use \texttt{code-davinci-002} given its excellent performance on \task{}. 
Table \ref{tab:pwd_results} presents the results of both methods. We can see that \method{} achieves competitive performance compared to the supervised \axcell{}, highlighting the efficacy of our proposed approach. When we enhance \method{} with \axcell{}'s cell selection capabilities to create \method{}+, it outperforms \axcell{}, demonstrating the promising potential of combining these two approaches.

\end{document}